\documentclass{article}
\PassOptionsToPackage{numbers, compress}{natbib}
\usepackage[eandd, final]{neurips_2026}

\usepackage[utf8]{inputenc} %
\usepackage[T1]{fontenc}    %
\usepackage{iftex}
\ifXeTeX
  \usepackage{newunicodechar}
  \newunicodechar{–}{--}
  \newunicodechar{’}{'}
\fi
\usepackage{hyperref}       %
\usepackage{url}            %
\usepackage{booktabs}       %
\usepackage{amsfonts}       %
\usepackage{nicefrac}       %
\usepackage{microtype}      %
\usepackage{xcolor}         %
\usepackage{amsmath}
\usepackage{subcaption}
\usepackage{graphicx}
\usepackage{wrapfig}
\usepackage{hyperref}
\usepackage{url}
\usepackage{booktabs}
\usepackage{pifont}%
\usepackage{xcolor} %

\newcommand{\sysnamenospace}{AtariBench}
\newcommand{\sysname}{AtariBench }
\newcommand{\yq}[1]{{\color{black}#1}}

\title{AtariBench: Evaluating Spatiotemporal Reasoning of Vision-Language Models in Atari Gameplay}

\author{%
  Mingjia Huo\textsuperscript{1,*} \quad Yao Fu\textsuperscript{2,\dag} \quad Bo Chang\textsuperscript{2} \quad Yaqing Wang\textsuperscript{2} \\
  \textbf{Dawei Zhu\textsuperscript{3,*} \quad Dylan Zhang\textsuperscript{4,*} \quad Xinyang Yi\textsuperscript{2} \quad Lichan Hong\textsuperscript{2}} \\
  \textbf{Ed H. Chi\textsuperscript{2} \quad Hao Zhang\textsuperscript{1} \quad Jiaxi Tang\textsuperscript{2}} \\
  \textsuperscript{1}University of California San Diego \quad \textsuperscript{2}Google DeepMind \\
  \textsuperscript{3}Peking University \quad \textsuperscript{4}University of Illinois Urbana-Champaign \\
  \texttt{mhuo@ucsd.edu} \quad \texttt{jiaxit@google.com}
}

\begin{document}
\addtocontents{toc}{\protect\setcounter{tocdepth}{-1}}

\maketitle
{\renewcommand{\thefootnote}{*}\footnotetext{Work done as Student Researcher at Google DeepMind.}}
{\renewcommand{\thefootnote}{\dag}\footnotetext{Work done while at Google DeepMind.}}

\begin{abstract}
Vision-Language Models (VLMs) have achieved strong performance on static visual understanding benchmarks, yet their ability to act in dynamic environments remains less well understood. We introduce AtariBench, a controlled benchmark that evaluates VLMs across 20 Atari 2600 games, requiring models to make sequential decisions from high-temporal-resolution gameplay streams. By decoupling fast in-game dynamics from wall-clock inference speed, AtariBench focuses on whether models can perceive rapid visual state changes, reason over spatiotemporal dynamics, and ground actions in the current observation. Our experiments reveal substantial limitations in state-of-the-art VLMs: models often fail to infer object motion and direction, exhibit brittle sequential decision-making, and over-rely on textual interaction history despite newly appended visual feedback. We further introduce a separability analysis to identify which in-game scores provide reliable signals of model competence, and use the benchmark to study how memory design and prompting strategies affect performance. Overall, AtariBench exposes a persistent gap between passive visual perception and active decision-making, providing a lightweight and rigorous testbed for developing VLMs that can reason and act in dynamic environments.
\end{abstract}
\section{Introduction}

\yq{Despite the remarkable proficiency of Vision-Language Models (VLMs) in static scene understanding \citep{mmmu}, the prevalent evaluation paradigm heavily over-indexes on static or turn-based benchmarks. This static-centric focus obscures a critical capability gap: when scaling to dynamic environments, current models struggle to reliably process continuous visual dynamics into actionable decisions \citep{videobench, mvbench, pca-bench, motionbench}. To bridge this gap, VLMs must evolve along three core dimensions: (1) \textit{Spatiotemporal Physics}, the intuitive grasp of physical dynamics such as object trajectory, distance, and velocity; (2) \textit{Sequential Decision-Making}, the capability to synthesize continuous visual streams into coherent, forward-looking strategies; and (3) \textit{Visual Grounding}, the ability to anchor decisions in the present, continuously updating visual state rather than relying on historical text or static priors.}

\yq{To systematically evaluate these interactive capabilities, we propose \sysnamenospace, which evaluates VLMs across 20 diverse Atari 2600 games.\footnote{Project page: \url{https://mignonjia.github.io/AtariBench/}.}
The Atari environment \citep{ale} has long served as a canonical testbed for agent-based AI \citep{atari_nature, badia2020agent57, go_explore}; in this work, we repurpose this classic platform as a diagnostic sandbox for modern VLMs. In our setting, models are tasked with making sequential decisions directly from high-temporal-resolution gameplay streams. By decoupling the fast-paced in-game dynamics from wall-clock model inference speed, we force models to process rapid state transitions and react sequentially, directly challenging their spatiotemporal reasoning. Our approach marks a deliberate departure from passive video-understanding benchmarks \citep{actionatlas, Perception_test_deepmind}. Furthermore, while other vision-to-action frameworks exist, their open-ended strategic and visual complexities can obscure the foundational reasoning failures we aim to isolate \citep{minedojo, videogamebench}, or they focus primarily on image-based planning rather than continuous video dynamics \citep{embodiedbench}. \sysnamenospace, in contrast, leverages constrained 2D physics as controlled environments to specifically measure these foundational abilities.}

\yq{Our experiments with \sysname reveal a cascade of critical failures in state-of-the-art models, highlighting the capability gaps identified above. First, we observe a pervasive weakness in motion comprehension: models consistently struggle to infer object movement and direction from sequential visual inputs, leading directly to poor in-game choices. Second, models suffer from long-context failures. Our qualitative analysis reveals that as the interaction history grows, models frequently collapse into pathological action loops, endlessly repeating past sequences regardless of newly appended visual feedback.
Third, text has mixed effects: textual outcome feedback can help models recover from missed changes in game state, while reliance on stale textual history can also lead to failures. Despite image tokens comprising the vast majority of the prompt, some decisions follow stale textual cues that conflict with the current frames, especially among weaker models.}

\yq{The \sysname framework offers several practical advantages for rigorous model evaluation. It provides a lightweight yet robust sandbox for rapid algorithmic iteration. While its simple but diverse game rules enable evaluation across a wide spectrum of cognitive tasks \citep{revisiting_ale, continuous_ale}, we go beyond raw scores by introducing a rigorous separability analysis. This ensures that the native in-game metrics reliably distinguish the evaluated models rather than reflecting random noise. Crucially, the framework also facilitates nuanced investigations into inference-time behaviors, allowing us to systematically evaluate how context history length and different prompted reasoning strategies impact empirical performance. Ultimately, by systematically exposing the gap between static perception and dynamic execution, \sysname establishes a rigorous baseline for evolving VLMs from passive observers into active agents.
}

\section{Related Work}

\paragraph{Game-Based Benchmarks for Language Models.}

Recent years have seen a surge of work on game-based benchmarks from diverse genres, including grid-based puzzles \citep{nasir2024gametraversalbenchmark}, simulation and resource-management strategy such as Factorio \citep{hopkins2025factorio}, turn-based role-playing with strategic combat such as Pokémon \citep{pokemon}, and narrative reasoning tasks such as murder-mystery games \citep{xie2024whodunitbench}. 
Others evaluate natural language reasoning through text-based or conversational games \citep{gamebench,textgames,gameeval,gamearena}, lacking visual understanding. Multimodal gaming benchmarks such as BALROG \citep{balrog}, LMAct \citep{lmact}, and LMGame-Bench \citep{lmgamebench} assess planning using image inputs, but their tasks are dominated by grid-based games that do not directly probe fine-grained spatial–temporal reasoning (e.g., movement trajectories, speed, or continuous dynamics) as found in Atari. VideoGameBench \citep{videogamebench} evaluates VLMs on AAA games, but most models make little progress due to the extreme complexity of these games, limiting its ability to separate models. Cradle \citep{Cradle} builds computer-use agents for such games but relies on modules like object detectors, making it hard to isolate VLM perception. In contrast, Atari games provide a lightweight testbed with simple rules and visuals, sufficient to probe spatial–temporal reasoning.
Atari-GPT \citep{atari_gpt} explores the use of VLMs on Atari environments, demonstrating initial feasibility. Our work, however, significantly extends this by constructing a systematic and reproducible benchmark designed for rigorous evaluation of VLM spatial–temporal reasoning. This involves a diverse set of games targeting motion perception, a standardized agentic harness, and repeated trials to ensure statistical reliability, thereby moving beyond a proof-of-concept to a robust framework.

\paragraph{Video Understanding, Action Recognition and Decision Making.}
General video benchmarks \citep{mvbench,videobench,Perception_test_deepmind} and fine-grained action suites \citep{actionatlas,motionbench} evaluate temporal dependencies and human motion comprehension primarily through video-based QA. VideoReasonBench \citep{videoreasonbench} challenges models with vision-centric tracking of altered states over time, while RIVER \citep{riverbench} evaluates timely responses to streaming videos. These works test video understanding without requiring control of the depicted environment. Existing action-oriented VLM benchmarks, such as EmbodiedBench \citep{embodiedbench} and PCA-Bench \citep{pca-bench}, also focus heavily on static, image-based decision making. AtariBench bridges this gap by requiring active, closed-loop control in dynamic, fast-paced interactive environments.

\section{Method}
\label{sec:method}

\begin{figure}
    \centering
    \includegraphics[width=\linewidth]{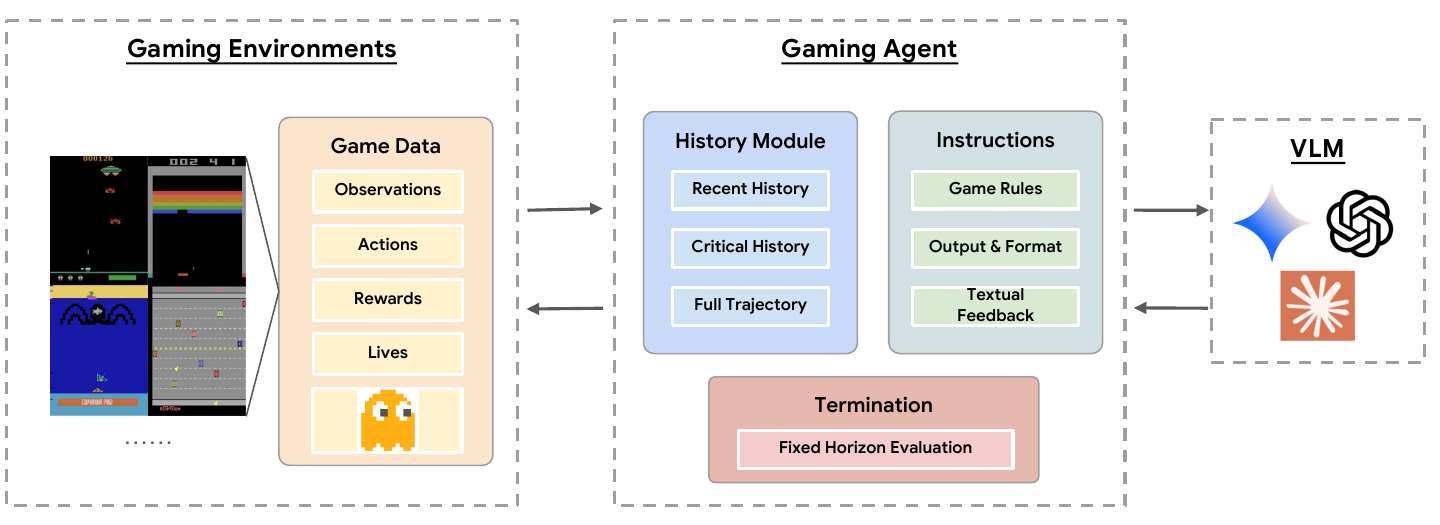}
    \caption{Workflow of AtariBench. At each turn, the harness assembles game instructions, interaction history, and the current frame into a prompt; the VLM returns a thought and an action sequence, which is executed in ALE, and the resulting frames and feedback are added to the history.}
    \label{fig:workflow}
\end{figure}

AtariBench is a benchmark for evaluating VLMs as game-playing agents in visually dynamic Atari game environments. We choose 20 fast-paced Atari games that cover diverse genres and goals, thereby probing a broad range of model capabilities. A standardized gaming-agent harness manages instructions, memory, action execution, and termination, enabling closed-loop interaction between VLM decisions and environment feedback.

\subsection{Arcade Learning Environment}

We implement AtariBench on top of the Arcade Learning Environment (ALE)~\citep{ale} via Gymnasium~\citep{gymnasium,openai_gym}. ALE provides a common control interface across Atari 2600 games, defining the action space, reward signal, observations, and termination conditions for each game. Each gaming environment exposes a standardized step API, and a low-level transition is formalized as:
\begin{verbatim}
obs, reward, terminated, truncated, info = env.step(action_id)
\end{verbatim}
Concretely, each interaction \texttt{step} takes an integer \texttt{action\_id} and returns an RGB observation frame (\texttt{obs}), a scalar reward (\texttt{reward}), termination flags (\texttt{terminated}, \texttt{truncated}), and environment metadata (\texttt{info}). This unified interface allows AtariBench to evaluate games under a single agent protocol. Appendix~\ref{appendix:ethics_release} discusses licensing, game assets, and our release policy.

\begin{wrapfigure}[12]{r}{0.5\textwidth}
    \centering
    \includegraphics[width=0.50\textwidth]{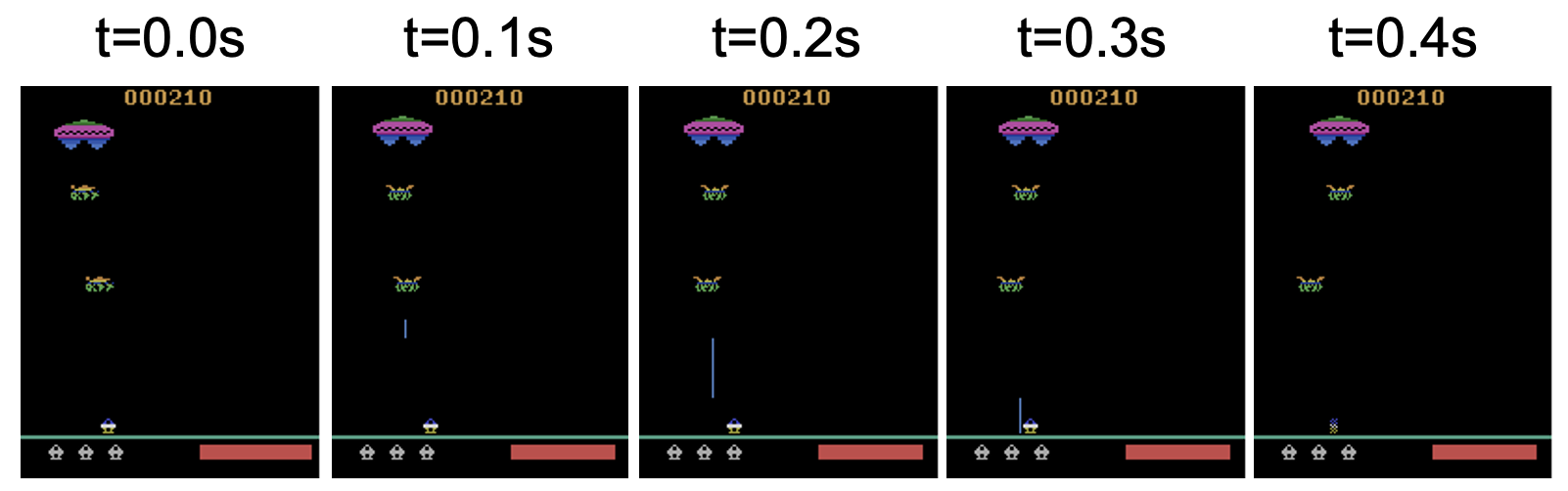}
    \caption{Fast temporal dynamics in \textit{Assault}. The enemy projectile appears at $t=0.1$s and already hits the player at $t=0.3$s, motivating high temporal resolution and 10~Hz control.}
    \label{fig:deciding_fps}
\end{wrapfigure}
\paragraph{Actions and Scores.} ALE defines 18 discrete actions, formed by the Cartesian product of horizontal movement (noop, left, right), vertical movement (noop, up, down), and a firing command (noop, fire). Some games support only a subset of these actions. For example, in \textit{Breakout}, the paddle moves only horizontally, and vertical actions are disabled. Meanwhile, the semantics of \texttt{fire} are game-dependent, such as starting a turn, shooting a fireball, or speeding up the movement. We provide the full action set and game-specific function of each action in the prompt when querying VLMs.

\paragraph{Control and Frame FPS.} We render frames at ALE's native resolution ($210\times160$). While ALE runs at 30~FPS, we execute actions at 10~Hz by repeating each chosen action for three frames. A model query can select several actions, and the environment pauses during inference. This retains fast object motion while keeping the token budget manageable for current VLMs. As illustrated in Fig.~\ref{fig:deciding_fps}, Atari dynamics can unfold within only a few tenths of a second: in \textit{Assault}, an enemy projectile appears at $t=0.1$s and reaches the player by roughly $t=0.3$s. Such examples motivate a sufficiently high observation and control frequency. We disable ALE sticky actions to remove stochasticity.

\paragraph{Fixed-Horizon Evaluation.} Each run is evaluated over a fixed 30-second window, with immediate reset after early game termination. Under the original life-based termination, a small survival instruction can change both episode length and reward in games where survival creates more scoring opportunities. The fixed horizon standardizes the interaction budget, reduces variance, and mitigates prompt-sensitive performance variation that makes it more difficult to accurately measure spatial-reasoning capabilities (Appendix~\ref{appendix:fix-horizon}).

\subsection{Game Choice and Taxonomy}
\label{sec:game_taxonomy}

ALE contains 100+ Atari 2600 games, from which we select 20 according to three principles.
First, each game should require \textbf{motion perception} rather than static
scene understanding alone, which rules out math, board, and card games.
Second, each game should be \textbf{human-interpretable}
under the fixed evaluation horizon.
After being given the game rules in advance,
a human player should be able to infer the controls and scoring
from a 30-second gameplay clip. This avoids games with ambiguous controls, sparse feedback, or opaque scoring rules that are difficult to learn within a short horizon. We manually played each selected game to ensure this.
Third, we prioritize coverage of \textbf{diverse} games, avoiding multiple games that are closely related variants of the same core mechanics and visuals; for example, we include \textit{PacMan} then not \textit{MsPacMan}.

Our selected games span shooters, sports, action games, and maze/navigation games. 
To make the benchmark analytically useful, we manually classify games along different aspects (Table~\ref{tab:game_choice}), including the primary skill (e.g., aiming, route planning, interception, scrolling navigation), number of different objectives (e.g., objectives that yield positive or negative points), action topology (horizontal, vertical, both axes, with optional fire/attack actions), reward density, visual load and complexity, and tempo sensitivity.
We include 11 games in the \textit{Shooter} category, defined as games in which the player must aim and shoot; despite this shared requirement of shooting, their mechanics and visuals actually vary substantially.
Moreover, most games require a high degree of temporal precision to achieve high scores.
This makes AtariBench a broad and challenging benchmark for spatiotemporal reasoning.

\begin{table}[t]
\centering
\small
\captionsetup{justification=raggedright,singlelinecheck=false}
\caption{AtariBench game taxonomy. \#Goals: number of distinct in-game goals. Action: A=fire/attack/throw, H=horizontal, V=vertical, B=both horizontal and vertical. 
Reward density and visual load are qualitative summaries. Tempo reflects the degree of timing precision required for competent play.}
\label{tab:game_choice}
\resizebox{\linewidth}{!}{
\begin{tabular}{lcccccc}
\toprule
\textbf{Game}  & \textbf{Primary Skill} & \textbf{\#Goals} & \textbf{Action} & \textbf{Reward Density} & \textbf{Visual Load} & \textbf{Tempo} \\
\midrule
\multicolumn{7}{c}{\it Shooter} \\
\midrule
AirRaid                & Aiming \& Avoidance         & 2 & H{+}A & Dense  & Low  & High \\
Assault                & Aiming \& Avoidance         & 3 & H{+}A & Dense  & Low  & High \\
BeamRider              & Aiming \& Avoidance         & 2 & H{+}A & Dense  & High & High \\
DemonAttack            & Aiming \& Avoidance         & 2 & H{+}A & Dense  & Low  & High \\
LaserGates             & Multi-task Navigation       & 2 & B{+}A & Dense  & High & High \\
NameThisGame           & Precision Interception      & 4 & H{+}A & Dense  & Low  & High \\
Phoenix                & Aiming \& Avoidance         & 2 & H{+}A & Dense  & Low  & High \\
RiverRaid              & Scrolling Navigation        & 3 & H{+}A & Dense  & High & High \\
Seaquest               & Multi-task Navigation       & 3 & B{+}A & Dense  & High & High \\
TimePilot              & 360-degree Navigation       & 2 & B{+}A & Sparse & High & High \\
Robotank               & First-person Navigation     & 2 & B{+}A & Dense  & High & High \\
\midrule
\multicolumn{7}{c}{\it Sports} \\
\midrule
Boxing                 & Rhythmic Combat             & 2 & B{+}A & Dense  & Low  & High \\
FishingDerby           & Depth Targeting             & 1 & B{+}A & Dense  & High & High \\
Tennis        & Predictive Timing           & 1 & B     & Sparse & Low  & Med/High \\
\midrule
\multicolumn{7}{c}{\it Action} \\
\midrule
Breakout      & Rebound Control             & 1 & H     & Dense  & Low  & Med/High \\
Freeway                & Gap Identification          & 2 & V     & Sparse & High & High \\
Gopher                 & Reactive Pathing            & 2 & H{+}A & Dense  & High & High \\
JourneyEscape          & Scrolling Navigation        & 1 & B     & Dense  & High & High \\
\midrule
\multicolumn{7}{c}{\it Maze} \\
\midrule
PacMan        & Route Planning              & 2 & B     & Dense  & High & High \\
Qbert                  & Route Planning              & 1 & B     & Dense  & High & Low  \\
\bottomrule
\end{tabular}
}
\end{table}

\subsection{Harness Design}

The harness defines the benchmark-side protocol through which VLMs interact
with Atari environments.
We intentionally use a simple yet necessary harness with only the components
needed for closed-loop gameplay:
task instructions, trajectory memory, environment feedback,
output parsing, and termination handling.
We do not optimize complex agent designs such as external planners or
tool-use modules.
While stronger harnesses may improve absolute scores, our goal is to compare
VLMs under the same controlled interaction interface.

At each decision point, the harness constructs a multimodal prompt with
the following components:
(i) a game-specific instruction block describing the objective and action
semantics,
(ii) historical context and the current frame derived from the current run,
and (iii) the VLM output schema.
The prompt template is shared across games, while game-specific instructions
provide per-game guidelines
(see Fig.~\ref{fig:breakout_prompt} for an example). We next describe the harness components in detail, including memory, environment feedback, and the VLM output schema.

\subsubsection{Memory}

To support sequential decision making under partial observability, we augment the current observation with lightweight memory derived from the agent's own trajectory. In the default \texttt{structured\_history} mode, each prompt contains the current frame together with two curated forms of episodic context: a short recent-history buffer and a sparse archive of reward-relevant events. The benchmark also supports an \texttt{append\_only} transcript mode, but our main harness uses the structured variant because it keeps context compact while preserving the most decision-relevant events.

\paragraph{Structured-History Mode.} Our memory harness contains two components. First, we include a \emph{recent-history buffer} consisting of the most recent $K$ turn-level clips, with $K=3$ by default. Second, we include a \emph{reward-history buffer} consisting of the most recent $R$ clips that produced non-zero reward, with $R=3$ by default. This yields around $5k$ input tokens per query. Life-loss events are also inserted into this reward-relevant memory even when the scalar reward is zero, because losing a life is an important gameplay failure that changes gameplay status and models' decisions.

Each clip corresponds to one completed model turn and stores: (i) the starting frame, (ii) the list of actions proposed by the model, and (iii) the resulting frames in action order, with one frame paired with each executed action. This representation provides the model with compact temporal evidence about what it recently did, what happened next, and which earlier situations were associated with success or failure. The recent-history buffer emphasizes short-horizon continuity, while the reward-history buffer highlights sparse but informative events that help the model infer game mechanics and effective strategies.

\paragraph{Append-Only Mode.} AtariBench also supports an alternative \texttt{append\_only} prompt mode. Rather than organizing past interaction into separate \texttt{recent-history} and \texttt{reward-history} blocks, this mode keeps the full trajectory just like a typical human-chatbot interaction. 
The initial user turn contains the game instructions, the time-budget rule, and the first frame. Each later assistant turn is the model's response, and each following user turn contains the resulting game frames with reward or life-loss feedback. This can lead to more than $100k$ input tokens per query.

We use this \texttt{append\_only} mode as an optional choice, while \texttt{structured\_history} remains our default. This is because we observe repetitive action sequences under \texttt{append\_only} prompting in evaluated models (see Sec.~\ref{sec:long_context_problem}).

\subsubsection{Textual Reward Feedback}

In addition to image frames, AtariBench exposes environment rewards through simple textual feedback. After each executed action, the pipeline emits short deterministic annotations whenever a salient event occurs. \textit{Positive or negative rewards} are verbalized as ``Feedback: You received a positive/negative score!'', and \textit{life-loss events} are described explicitly as ``Feedback: A life was lost and a new life has begun.'' If no salient event occurs, no extra feedback text is added.

Textual feedback makes rewards and failure events explicit since it provides clear signals about the consequences of the model's actions. For instance, in Breakout, after the model loses the ball, the screen can enter a reset state with no visible ball. If the model fails to infer from the frame that it must press \texttt{start} to serve the next ball, the game can stall after the initial loss; textual feedback makes this reset explicit, and we discuss its effect in detail in Sec.~\ref{sec:qualitative_analysis}.

\subsubsection{Model Output Schema}
Each game exposes a normalized textual action vocabulary derived from the ALE action set used by that game. The model is required to produce a short free-form \textit{thought} followed by 1-10 \textit{actions}. The harness parses the output, normalizes action names, validates them against the game-specific action map, and executes the resulting sequence in order. Invalid outputs are handled conservatively by falling back to a \texttt{no-op} action, which keeps evaluation robust while preserving the raw response for later analysis.
Variable-length sequences permit short action sequences at critical frames while balancing evaluation cost (Appendix~\ref{appendix:action_chunk}).

\section{Experiments}
\label{sec:experiments}
This section presents the empirical evaluation of AtariBench. We first compare state-of-the-art VLMs and a random-action baseline across AtariBench games. We then analyze game separability to ensure that the retained games can distinguish models. We further present qualitative case studies of recurring failures. Finally, we diagnose perception-to-action skills with controlled QA questions; supplementary gameplay-level failure attribution using VLM-as-a-judge is provided in Appendix~\ref{appendix:vlm_judge}.

\subsection{Results}

\begin{figure}[t]
    \centering
    \includegraphics[width=\linewidth]{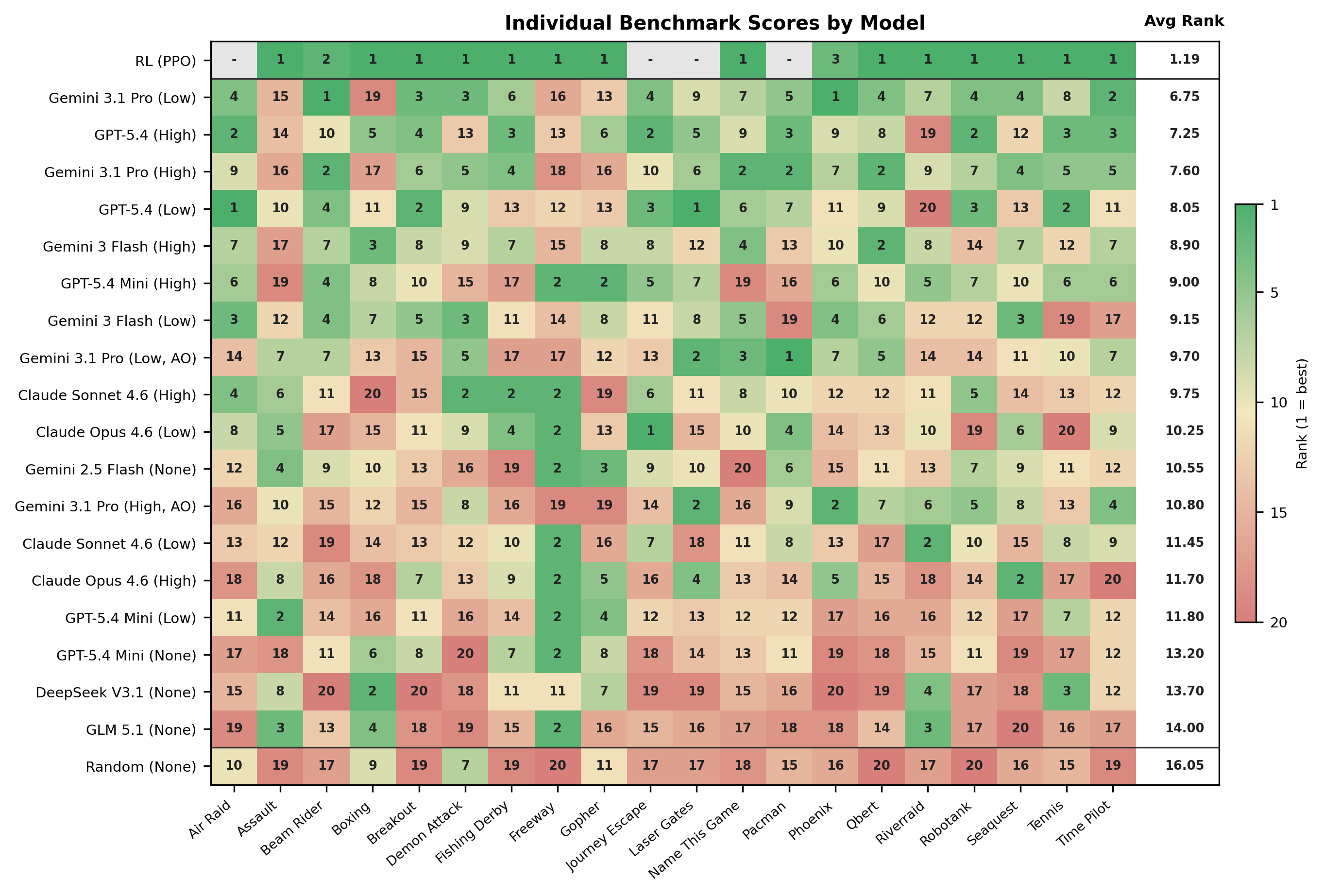}
    \caption{
    Model performance ranks across the 20 AtariBench games used in the main analysis, including the standard structured-history settings and the Gemini 3.1 Pro append-only (AO) settings. Game-specific RL agents and a random-action baseline are also included. Rows are ordered by average per-game rank; raw reward values are reported in Appendix~\ref{appendix:raw_scores}.}
    \label{fig:main}
\end{figure}

Fig.~\ref{fig:main} reports the main evaluation across 20 games, comparing Gemini 3.1 Pro, Gemini 3 Flash, Gemini 2.5 Flash, GPT-5.4, GPT-5.4 Mini, Claude Opus 4.6, Claude Sonnet 4.6, DeepSeek V3.1, GLM 5.1, and a random-action baseline. We also vary the thinking effort levels for eligible models. The figure also includes Gemini 3.1 Pro append-only variants, marked as ``AO''. For each game, we evaluate each model setting over ten 30-second closed-loop runs using ALE seeds 0--9 under the same execution harness, with sticky actions disabled. Overall, Gemini 3.1 Pro and GPT-5.4 occupy the strongest average-rank positions, while the random baseline remains consistently near the bottom, indicating that the benchmark rewards learned visual control rather than arbitrary action sampling. We evaluate PPO-based game-specific RL agents from Sample Factory~\citep{sample_factory} on the 16 games with matching public checkpoints. They rank first on 14 of these games, highlighting a substantial performance gap between state-of-the-art VLMs and specialized game policies; Appendix~\ref{appendix:rl_baseline} gives the evaluation details.

\subsection{Game Separability Analysis}

Here we assess how well our selected Atari games distinguish among models.
Even with an interpretable objective and a valid reward function, a game can be a weak benchmark if no model beats random play or if scores cluster into the same narrow range. To ensure each game yields informative signal, we apply three complementary criteria that test model separability.

First, the best VLM setting must outperform random play, with a positive mean-reward gap, pooled-SD effect size $d\ge0.8$, and $q<0.05$ from a two-group ANOVA \cite{fisher1934statistical}. This criterion directly asks whether any evaluated model can exploit the game structure better than random play. Second, the top and bottom quarters of non-random settings (each $\lceil N/4\rceil$ settings) must have a positive mean gap and $d\ge0.8$. This top--bottom contrast is useful when only a subset of models discovers the right behavior while many others remain similarly weak. Third, a one-way ANOVA \cite{oneway_anova_wikipedia} across all settings, including random play, must give $q<0.05$ and $\eta^2\ge0.14$. The ANOVA tests whether mean rewards differ across settings, while $\eta^2$ measures the fraction of observed score variance attributable to between-setting differences. For the first and third criteria, we apply Benjamini--Hochberg correction \cite{FDR} separately across the 20 games.

\begin{figure}[t]
    \centering
    \includegraphics[width=0.92\linewidth]{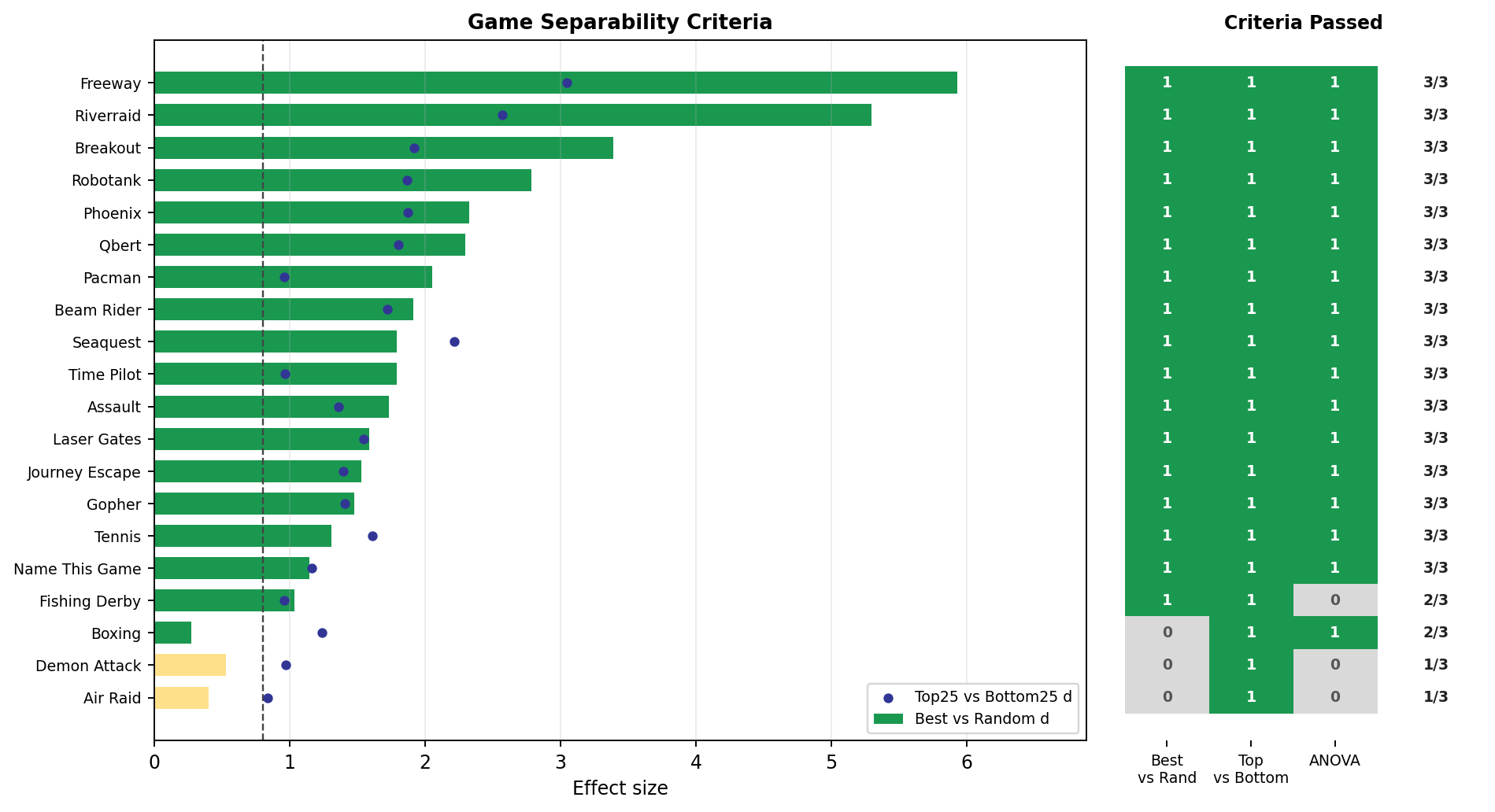}
    \caption{Game separability analysis across the 20 AtariBench games. Bars show the best-model-vs-random effect size, points show the top-quartile-vs-bottom-quartile effect size, and the right-hand matrix marks whether each of the three criteria passes. Games are strongly separable when at least two criteria pass and weakly separable when exactly one passes.}
    \label{fig:game_usefulness}
\end{figure}

Figure~\ref{fig:game_usefulness} summarizes the resulting game-level labels using a simple pass-count rule. A game is \emph{strongly separable} when at least two of the three criteria pass, \emph{weakly separable} when exactly one criterion passes, and not separable when none pass. Under this criterion, all 20 games retained in AtariBench are separable: 18 are strongly separable, and the remaining 2 are weakly separable. This result separates benchmark validity from broad model disagreement while keeping the decision rule transparent.

\subsection{Qualitative Analysis}
\label{sec:qualitative_analysis}

\paragraph{Visual Perception.}
AtariBench exposes failures to detect visible objects. In the controlled diagnostics, perception accuracy ranges from 61.0\% to 100.0\% across model settings (Table~\ref{tab:mini_eval}).  In Name This Game (Fig.~\ref{fig:object_omission_perception}), GPT-5.4 (\texttt{high}) states that no shark is visible, although a white shark appears in the current frame, and its plan anticipates a shark appearing rather than acknowledging the one already present (Appendix~\ref{appendix:qualitative_failure_cases}). Models can also misidentify objects, such as mistaking the controlled aircraft for an enemy in Time Pilot (Appendix~\ref{appendix:vlm_judge}).

\paragraph{Temporal Misgrounding.}
Models can describe a historical spatial relation as if it were current.
In Freeway, GPT-5.4 (\texttt{high}) describes a blue car as still to the
chicken's right, although it has already crossed to the left
(Fig.~\ref{fig:temporal_misgrounding}). The claim matches earlier frames
provided in the same prompt, consistent with a failure to update the spatial
relation at the current time.
Appendix~\ref{appendix:spatial_history_controls} further examines this failure by comparing the model's spatial understanding when given the complete history versus only the current frame: on spatial-relation failure cases, spatial errors persist in 39.3\% of complete-history re-queries but in only 10.7\% of current-frame-only re-queries (exact McNemar $p=0.0386$). This suggests that interaction history can interfere with grounding the current state, so additional interaction history is not necessarily beneficial.

\paragraph{Spatiotemporal Understanding.}

Motion comprehension is a major challenge, as indicated by the diagnostic questions in Sec.~\ref{sec:multi_choice_eval} and the VLM-judge analysis in Appendix~\ref{appendix:vlm_judge}. In a Breakout example
(Fig.~\ref{fig:breakout_motion}), the inferred movement direction contradicts
the ground-truth trajectory. This error can propagate into planning without
being repaired by additional reasoning
(Appendix~\ref{appendix:reasoning_propagation}), compromising object tracking
and action selection in dynamic scenes. This is consistent with the benchmark results: higher reasoning effort does not consistently improve scores and can sometimes reduce performance.

\paragraph{Text-History Overreliance.}

Textual outcome feedback can both help and hurt model performance. It can help weaker models identify critical gameplay events, but it can also hurt performance when models rely on outdated feedback instead of current visual observations.
In Breakout, for example, a life-loss notice helps Gemini 2.5 Flash serve the next ball with \texttt{start}, and our ablation analysis finds a significant score benefit for this model on Breakout (Appendix~\ref{appendix:feedback_ablation}). However, earlier feedback can become stale as the game evolves. In the Assault example in Fig.~\ref{fig:assault_text_overreliance}, GPT-5.4 Mini interprets an earlier positive-score notice as evidence that recent upward shots have been hitting, so it continues firing upward without accounting for the current game state.

To measure the prevalence of these failures, we use a VLM judge and find that text-history overreliance accounts for 22.4\% of observable failures in the two weaker model settings and 4.9\% in the two stronger settings (Appendix~\ref{appendix:text_history_audit}). This suggests that the stronger models make more effective use of current visual evidence.

\paragraph{Long-Context Problem.}\label{sec:long_context_problem}

\begin{figure}[t]
    \centering
    \includegraphics[width=\linewidth]{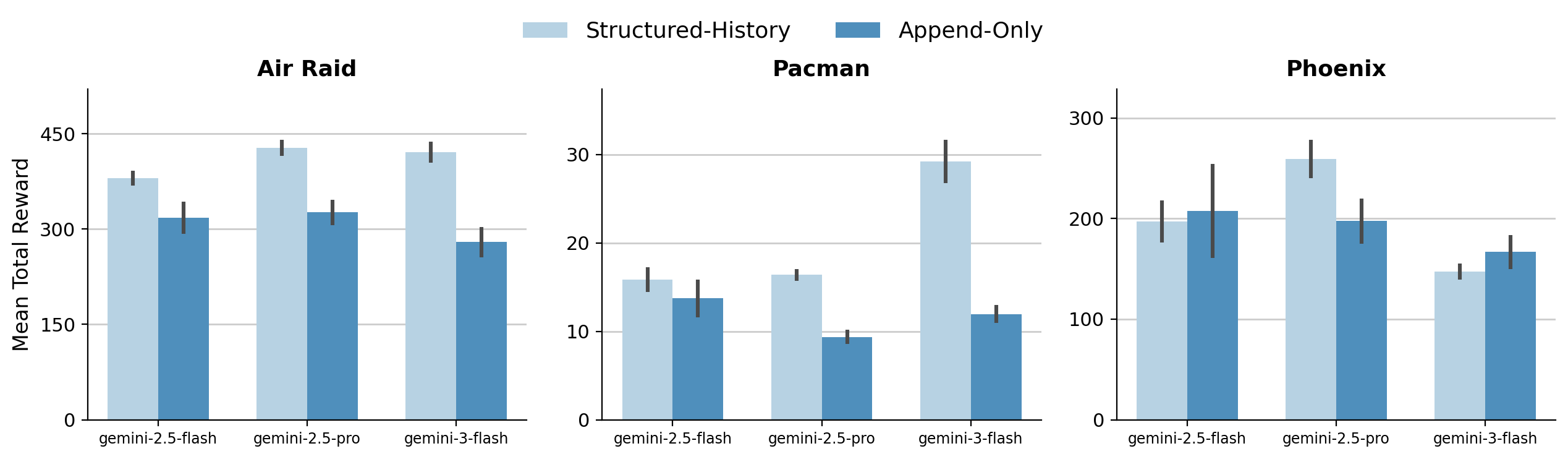}
    \caption{Append-only versus structured-history performance on three representative games for Gemini 2.5 Flash, Gemini 2.5 Pro, and Gemini 3 Flash. Bars show mean total reward, and error bars show standard error over runs. All models use default thinking.}
    \label{fig:append_only_non_gemini3pro}
\end{figure}

We also evaluate the \texttt{append\_only} prompting format, which more closely follows the long-context chat interface used by modern LLMs. This makes the input as long as 100k tokens, because the entire 30-second history is included in one call. For Gemini 2.5 Flash, Gemini 2.5 Pro, and Gemini 3 Flash, the comparison on Air Raid, Pacman, and Phoenix shows that appending the full interaction history usually does not improve over the structured-history baseline (Fig.~\ref{fig:append_only_non_gemini3pro}). These results suggest that append-only context does not consistently improve gameplay rewards in the evaluated settings.

One failure mode is action stagnation. For Gemini 2.5 Flash (\texttt{off}) on Breakout, noop-only decisions account for $48.8\%$ of turns under append-only prompting versus $0.7\%$ under structured-history prompting, measured over the latter half of decision turns within each run and averaged across ten runs per history setting. Fig.~\ref{fig:append_only_failure_example} shows a concrete Breakout trace where the model repeatedly emits only \texttt{noop} actions despite receiving new frames and feedback.

\begin{table}[t!]
\centering
\footnotesize
\caption{Model accuracy on 60 manually curated diagnostic questions, with five queries per question.}
\setlength{\tabcolsep}{4pt}
\begin{tabular}{lcccc}
\toprule
\textbf{Model (reasoning)} & \textbf{Perception} & \textbf{Motion} &
\textbf{Decision} & \textbf{Overall} \\
\midrule
Gemini 2.5 Flash (off) & 88.0\% & 51.0\% & 48.0\% & 62.3\% \\
Gemini 2.5 Flash (on) & 85.0\% & 47.0\% & 65.0\% & 65.7\% \\
Gemini 2.5 Pro (default) & 93.0\% & 54.0\% & 79.0\% & 75.3\% \\
Gemini 3 Flash (low) & 98.0\% & 64.0\% & 88.0\% & 83.3\% \\
Gemini 3 Flash (high) & 99.0\% & 63.0\% & 82.0\% & 81.3\% \\
Gemini 3.1 Pro (low) & \textbf{100.0\%} & 69.0\% & 82.0\% & 83.7\% \\
Gemini 3.1 Pro (high) & \textbf{100.0\%} & 60.0\% & 80.0\% & 80.0\% \\
GPT-5.4 (low) & 91.0\% & \textbf{74.0\%} & \textbf{94.0\%} & \textbf{86.3\%} \\
GPT-5.4 (high) & 91.0\% & 73.0\% & 83.0\% & 82.3\% \\
Claude Opus 4.6 (low) & 61.0\% & 65.0\% & 65.0\% & 63.7\% \\
Claude Opus 4.6 (high) & 65.0\% & 72.0\% & 67.0\% & 68.0\% \\
\bottomrule
\end{tabular}
\label{tab:mini_eval}
\end{table}

\subsection{Failure Modes Breakdown}\label{sec:failure_analysis}

\label{sec:multi_choice_eval}

To diagnose why strong VLMs still struggle in closed-loop Atari gameplay, we
construct a small, manually labeled diagnostic set.
We select ten games and curate 20 perception questions, 20 motion
questions, and 10 decision questions: two perception, two motion, and one
decision question per game. Each decision question is presented with a short
and a long prompt, yielding 60 diagnostic question variants in total.

Each sample consists of a short sequence of
frames selected from an actual benchmark trajectory, together with a small set
of closed-choice questions. For every question, each model is queried five
times and accuracy is computed over the repeated trials. This design isolates
specific perception-to-action skills while keeping the answer space easy to
parse automatically. Appendix~\ref{appendix:judge_calibration} provides the diagnostic prompts and calibration details.

The diagnostic questions are organized into four fine-grained capability
groups:

\begin{enumerate}
    \item \textbf{Perception: detection.} These questions test whether the
    model can identify salient objects in a single frame, such as counting fish
    or detecting the object relevant to the next action.
    \item \textbf{Perception: spatial reasoning.} These questions test whether
    the model can compare positions of visible objects in a single frame, such
    as whether the player's controllable object is currently to the left or
    right of a target.
    \item \textbf{Motion comprehension.} These questions use a timestamped
    frame sequence and ask for object or controller motion, such as whether a
    target is moving left or right, whether the player's line is moving
    up-left, down-left, or still, or which tile changes color after a hop.
    \item \textbf{Decision making.} These questions ask for the next action
    implied by the state. We evaluate both a short decision prompt and a long
    decision prompt. The long version includes the game's action-planning
    guidelines before asking the same closed-choice decision question.
\end{enumerate}

Table~\ref{tab:mini_eval} reports accuracy over five queries per question.
Across all VLMs, perception accuracy ranges from 61.0\% to 100.0\%,
motion accuracy from 47.0\% to 74.0\%, and decision accuracy from
48.0\% to 94.0\%. Errors in these closed-choice tasks expose weaknesses even when the action space and intended goal are specified. Most VLMs score higher on perception than on
motion, while higher reasoning effort does not consistently improve overall accuracy.

\paragraph{VLM-judge.} Attributing a failure in a full gameplay turn is more complex than evaluating
an isolated diagnostic question: perception, temporal grounding, action
selection, and subsequent game dynamics can all contribute to the outcome.
For example, an apparent relative-position error may reflect either a spatial error or a temporal misgrounding that relies on a historical frame rather than the current observation.
We therefore focus on controlled QA in the main analysis and provide the gameplay-level VLM-judge analysis in Appendix~\ref{appendix:vlm_judge} for reference. Among the judged failures, motion and trajectory-prediction failures are the most frequent, followed by decision and control, spatial relations, and visual perception, while exploration and rule-understanding failures are rare. A follow-up judge analysis further suggests that some spatial-relation errors stem from temporal misgrounding (Appendix~\ref{appendix:spatial_history_controls}). This provides a larger-scale, gameplay-level diagnosis compared with the above QA set.

\section{Conclusion}
We introduced AtariBench, a spatiotemporal gameplay benchmark built on the Arcade Learning Environment to evaluate the physical reasoning and closed-loop decision-making capabilities of modern VLMs. Our experiments show that even state-of-the-art models struggle to convert fast visual dynamics into reliable actions, often failing to infer object motion, update decisions from current frames, or use long interaction histories effectively. Diagnostic analysis identifies spatiotemporal reasoning as the most frequent failure category in the audited sample, alongside perception and control errors. Overall, AtariBench reveals a persistent gap between passive visual understanding and active control, highlighting the need for future VLMs to be trained and evaluated in dynamic interactive environments.

\begin{ack}
We thank Google DeepMind for supporting this work. The authors have no additional competing interests to declare beyond the affiliations and employment relationships noted on the first page.
\end{ack}

\bibliography{ref}
\bibliographystyle{splncs}

\newpage
\appendix
\addtocontents{toc}{\protect\setcounter{tocdepth}{2}}
\begingroup
\renewcommand{\contentsname}{Appendix Contents}
\hypersetup{hidelinks}
\tableofcontents
\endgroup
\clearpage

\section{Evaluation Protocol and Prompt}
\label{appendix:evaluation_protocol}

\subsection{Fixed-Horizon Evaluation and Strategy Sensitivity}
\label{appendix:fix-horizon}

Original Atari episodes often terminate after a fixed number of lost lives. We do not use this game-over termination as the primary evaluation protocol because the resulting score can be strongly confounded by prompt-induced strategy. In some games, survival itself creates more scoring opportunities, so a small survival instruction can change both behavior and reward.

The old game-over protocol also has high per-run variability because the evaluation length is itself model- and strategy-dependent. Table~\ref{tab:termination_variance} compares the original non-fixed-horizon Gemini 2.5 Flash runs with the current fixed-horizon runs on DemonAttack and Phoenix, using only the matched old setting: no thinking, structured visual prompting, three history clips, and no reward examples. The old protocol has substantially larger per-run standard deviation on both games.

\begin{table}[h!]
\centering
\footnotesize
\caption{Gemini 2.5 Flash reward variability under original game-over termination versus the current fixed 30-second protocol. We report mean reward $\pm$ standard error and per-run standard deviation (SD).}
\begin{tabular}{l|l|ccc}
\toprule
Game & Protocol & Runs & Mean $\pm$ SE & SD \\
\midrule
DemonAttack & Game-over termination & 30 & $70.0 \pm 6.6$ & $36.3$ \\
DemonAttack & Fixed 30 seconds & 10 & $41.0 \pm 4.3$ & $13.7$ \\
Phoenix & Game-over termination & 30 & $171.3 \pm 19.5$ & $106.6$ \\
Phoenix & Fixed 30 seconds & 10 & $198.0 \pm 27.4$ & $86.6$ \\
\bottomrule
\end{tabular}
\label{tab:termination_variance}
\end{table}

As a diagnostic, we evaluated a Phoenix prompt variant that added the sentence: ``Important: You will lose a life if you are hit by the bombs. The game will end if you lose three lives. So avoiding being hit is the top priority!'' Under game-over termination, this prompt shifted Gemini 2.5 Flash toward more survival-focused behavior and increased its average reward from $177.9$ to $195.4$ (see Table~\ref{tab:termination_strategy_sensitivity}). This illustrates why AtariBench instead uses a fixed 30-second horizon with immediate reset after early termination: the benchmark measures closed-loop control over a standardized time budget rather than a model's prompt-sensitive risk preference.

\begin{table}[h!]
\centering
\footnotesize
\caption{Legacy strategy-sensitivity diagnostic under game-over termination. We report the number of reasoning traces classified as reward-seeking or survival-focused out of 100 sampled traces, along with the average reward at termination.}
\resizebox{\linewidth}{!}{%
\begin{tabular}{l|cccc|cccc}
\toprule
& \multicolumn{4}{c|}{Phoenix (neutral)}
& \multicolumn{4}{c}{Phoenix (survival prompt)} \\
\cmidrule(lr){2-5} \cmidrule(lr){6-9}
Model
& Reward & Survival & Neutral & Avg
& Reward & Survival & Neutral & Avg \\
& seeking & focused & & Reward
& seeking & focused & & Reward \\
\midrule
Gemini 2.5 Flash & 76 & 19 & 5 & 177.9 & 68 & 25 & 7 & 195.4 \\
Gemini 2.5 Pro   & 59 & 34 & 7 & 239.5 & 80 & 16 & 4 & 243.8 \\
\bottomrule
\end{tabular}
}
\label{tab:termination_strategy_sensitivity}
\end{table}

\subsection{Evaluation Prompt}
\label{appendix:prompts}

\begin{figure}[htbp]
    \centering
    \includegraphics[width=0.9\linewidth]{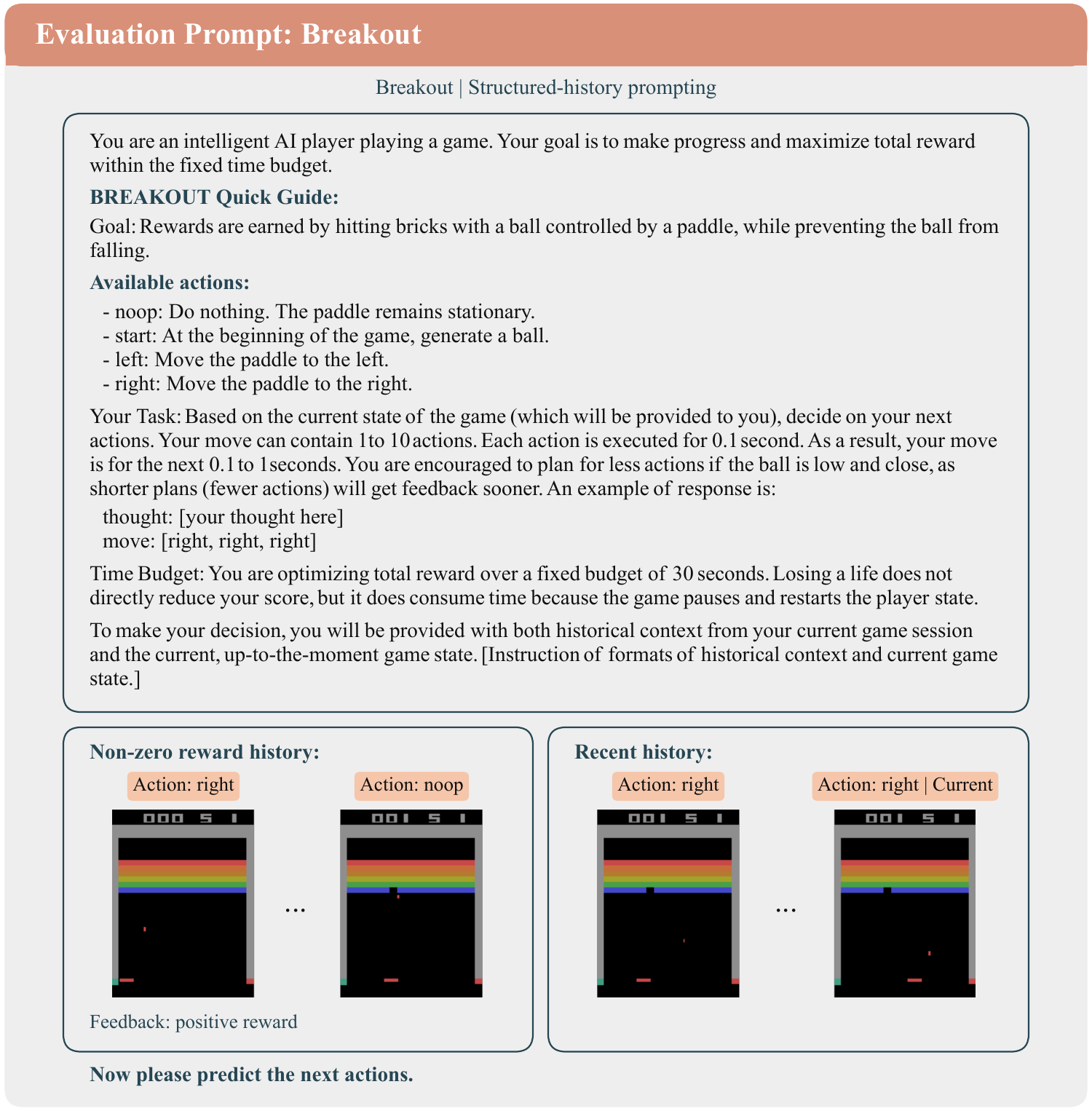}
    \caption{Example evaluation prompt for Breakout.}
    \label{fig:breakout_prompt}
\end{figure}

Fig.~\ref{fig:breakout_prompt} illustrates the evaluation prompt used in AtariBench, with Breakout as an example. The prompt combines game-specific instructions, recent and reward-relevant history, the current observation, and the required response format.

The Breakout prompt instantiates the harness described in Sec.~\ref{sec:method}. It begins with the game-specific instruction block, including the objective, available actions, action timing, and game-end condition. The prompt then provides the benchmark-managed memory used in the default \texttt{structured\_history} mode: non-zero-reward history, recent history, and the current game state. Each historical clip records the model's previous actions, the resulting frames, and any textual reward or life-loss feedback. Finally, the prompt fixes the output schema by requiring a free-form \texttt{thought} line followed by a \texttt{move} list, which the harness parses into ALE actions for closed-loop execution.

\section{Additional Results}
\label{appendix:additional_results}

\subsection{Raw Scores}
\label{appendix:raw_scores}

\begin{figure}[h!]
    \centering
    \includegraphics[width=\linewidth]{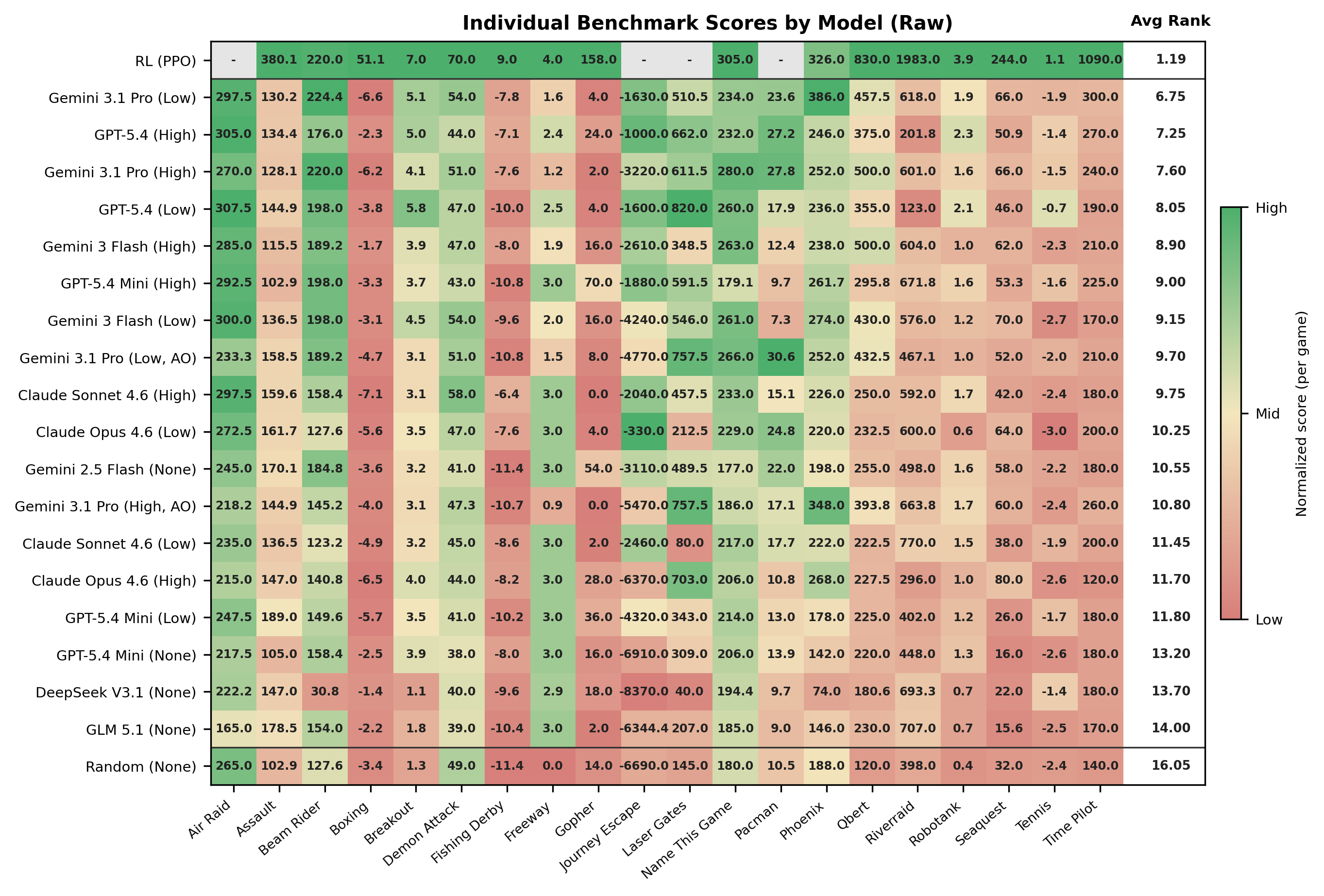}
    \caption{Raw model rewards across the 20 AtariBench games used in the main analysis, including the Gemini 3.1 Pro append-only settings shown in Fig.~\ref{fig:main}. The first row reports rewards for game-specific RL agents on 16 games, with ``-'' indicating unavailable results. Cell values preserve each game's raw reward scale, while colors are normalized within each game. Avg Rank is computed as in Fig.~\ref{fig:main}.}
    \label{fig:raw_scores}
\end{figure}

Fig.~\ref{fig:raw_scores} complements the main rank heatmap by preserving each game's raw reward scale. This view is useful for checking whether rank differences correspond to large reward gaps or to small differences on games with narrow score ranges.

\subsection{Game-Specific RL Baseline}
\label{appendix:rl_baseline}

We evaluate game-specific RL agents trained with APPO using Sample Factory~\citep{sample_factory} on 16 games shared with AtariBench. Each agent is evaluated using ten seeds, a 30-second gameplay duration, and no sticky actions. Scores are computed from raw ALE rewards before Sample Factory's reward clipping, matching the VLM scoring protocol. Figs.~\ref{fig:main} and~\ref{fig:raw_scores} include the RL results. These agents rank first on 14 of the 16 games, with an average rank of $1.19$. These results suggest that the state-of-the-art VLMs evaluated in AtariBench still lag behind specialized RL agents.

\textit{Why 16 shared games?} We selected AtariBench's 20 games independently using the motion-perception, short-horizon interpretability, and game-diversity criteria in Sec.~\ref{sec:game_taxonomy}. For the RL comparison, we use publicly released Atari checkpoints listed in the Sample Factory documentation.\footnote{\url{https://www.samplefactory.dev/09-environment-integrations/atari/}} Matching public checkpoints are available for 16 of our games. The overlap follows checkpoint availability and was not selected based on RL scores; it retains shooters, sports, action games, and maze navigation.

\section{Harness Ablations}

\subsection{Influence of Textual Feedback}
\label{appendix:feedback_ablation}

To test the influence of textual outcome feedback, we compare the standard
prompt with a variant that removes reward, life-loss, and reset notices. We
retain visual history, action labels, and the history-selection mechanism,
including the selection of reward-relevant clips. We select one game from each of the four genres listed in Table~\ref{tab:game_choice}: Breakout, Air Raid, Pacman, and Fishing Derby. We evaluate two models, Gemini 3.1 Pro (low) and Gemini 2.5 Flash (off), using ten seeds per setting.

\begin{table}[htbp]
\centering
\small
\caption{Textual feedback ablation. Scores are mean $\pm$ SE over ten seeds.}
\label{tab:feedback_ablation}
\begin{tabular}{@{}llrr@{}}
\toprule
Model & Game & Full feedback & No feedback \\
\midrule
Gemini 3.1 Pro & Breakout & $4.5\pm0.2$ & $4.7\pm0.3$ \\
& Air Raid & $225.0\pm23.9$ & $215.0\pm29.6$ \\
& Pacman & $31.6\pm5.4$ & $26.7\pm3.1$ \\
& Fishing Derby & $-8.4\pm1.8$ & $-8.6\pm1.4$ \\
\midrule
Gemini 2.5 Flash & Breakout & $2.3\pm0.3$ & $1.5\pm0.3$ \\
& Air Raid & $255.0\pm45.3$ & $282.5\pm30.1$ \\
& Pacman & $18.0\pm3.6$ & $18.0\pm2.6$ \\
& Fishing Derby & $-9.0\pm1.3$ & $-11.8\pm0.2$ \\
\bottomrule
\end{tabular}
\end{table}

Adding textual feedback significantly improves the mean scores of Gemini 2.5 Flash on Breakout and Fishing Derby. Significance is determined by whether the paired bootstrap 95\% CI excludes zero (10,000 resamples; no multiple-comparison correction). For Gemini 3.1 Pro, no statistically significant differences are observed across the four games, indicating no detectable benefit from textual feedback for the stronger model.

\begin{figure}[h!]
    \centering
    \includegraphics[width=0.9\linewidth]{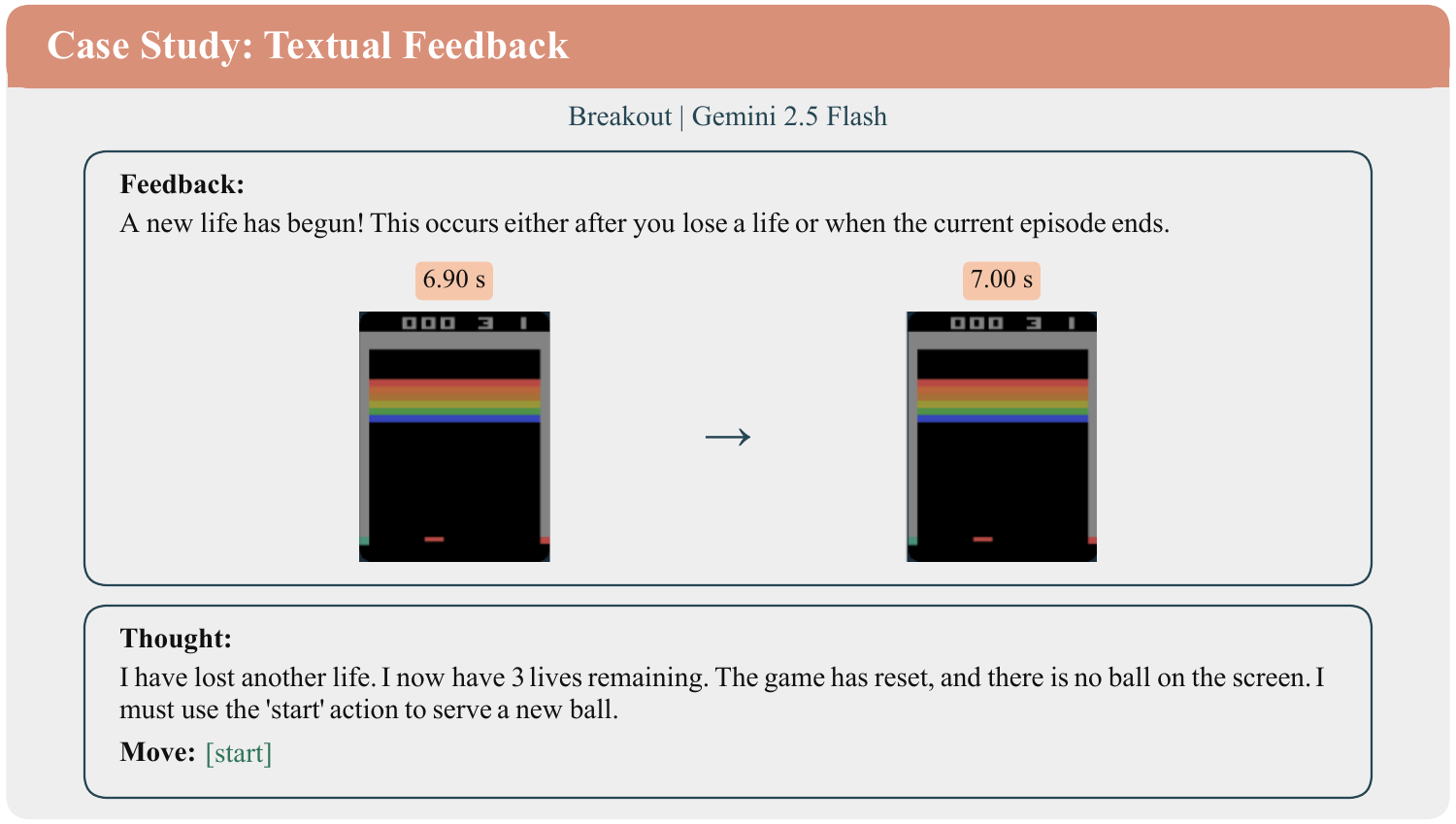}
    \caption{Breakout example with Gemini 2.5 Flash in which textual life-loss feedback makes the reset state explicit. After the harness reports that a new life has begun, the model selects \texttt{start}, which serves the next ball and resumes play. Without textual feedback, the model stalls and fails to press \texttt{start} to proceed.}
    \label{fig:text_feedback_breakout}
\end{figure}

By inspecting the reasoning traces of Gemini 2.5 Flash, we find that textual feedback helps the weaker model interpret the game state and choose appropriate actions. Without textual feedback, weaker models playing Breakout can stall after losing a ball if they fail to press \texttt{start} to serve another (Fig.~\ref{fig:text_feedback_breakout}). Such stalls can leave scores at only 0 or 1 over the 30-second window, making performance difficult to distinguish. These observations suggest that textual feedback can help weaker models make progress.

\subsection{Action-Chunk Length}
\label{appendix:action_chunk}

We vary the permitted number of actions per model query: exactly one,
one to five, or the default one to ten. We use the standard textual output format and
structured-history prompt for Gemini 3 Flash (low) and GPT-5.4 Mini
(none) on the same four games selected above.

\begin{table}[htbp]
\centering
\small
\caption{Reward under different allowed action-chunk lengths (mean $\pm$ SE
over ten seeds). Variable-length conditions permit the model to choose shorter
sequences at each turn.}
\label{tab:action_chunk}
\begin{tabular}{@{}llrrr@{}}
\toprule
Model & Game & 1 action & 1--5 actions & 1--10 actions \\
\midrule
Gemini 3 Flash & Air Raid & $295.0\pm33.3$ & $340.0\pm36.7$ & $300.0\pm31.0$ \\
& Breakout & $4.6\pm0.2$ & $4.5\pm0.2$ & $3.7\pm0.3$ \\
& Fishing Derby & $-6.6\pm2.1$ & $-8.4\pm1.7$ & $-7.2\pm1.4$ \\
& Pacman & $17.4\pm4.0$ & $13.6\pm3.1$ & $11.9\pm2.8$ \\
\midrule
GPT-5.4 Mini & Air Raid & $175.0\pm18.3$ & $265.0\pm39.8$ & $217.5\pm41.2$ \\
& Breakout & $3.4\pm0.2$ & $3.8\pm0.3$ & $3.9\pm0.3$ \\
& Fishing Derby & $-10.2\pm0.8$ & $-9.8\pm0.6$ & $-8.0\pm0.8$ \\
& Pacman & $12.4\pm1.5$ & $11.9\pm1.3$ & $13.9\pm1.8$ \\
\bottomrule
\end{tabular}
\end{table}

Seven of the eight model–game pairs show no significant difference between the default 1–10 range and shorter settings. Significance is determined by whether the bootstrap 95\% CI of the score difference excludes zero (10,000 resamples; paired by seed where seeds match, otherwise unpaired; no multiple-comparison correction). The only significant difference is observed for Gemini 3 Flash on Breakout, where one-action control scores higher than the default range ($4.6$ versus $3.7$), suggesting that finer-grained control can help in games that require precise timing. We retain the 1–10 action range because it allows the model to choose short action sequences when needed while balancing evaluation cost.

\section{Failure Analysis and Diagnostics}
\label{appendix:failure_details}

\subsection{Qualitative Failure Cases}
\label{appendix:qualitative_failure_cases}

We provide examples of the visual-perception, temporal-grounding, motion-reasoning, text-history, and long-context action-selection failures discussed in Sec.~\ref{sec:qualitative_analysis}.

\begin{figure}[tbp]
    \centering
    \includegraphics[width=0.9\linewidth]{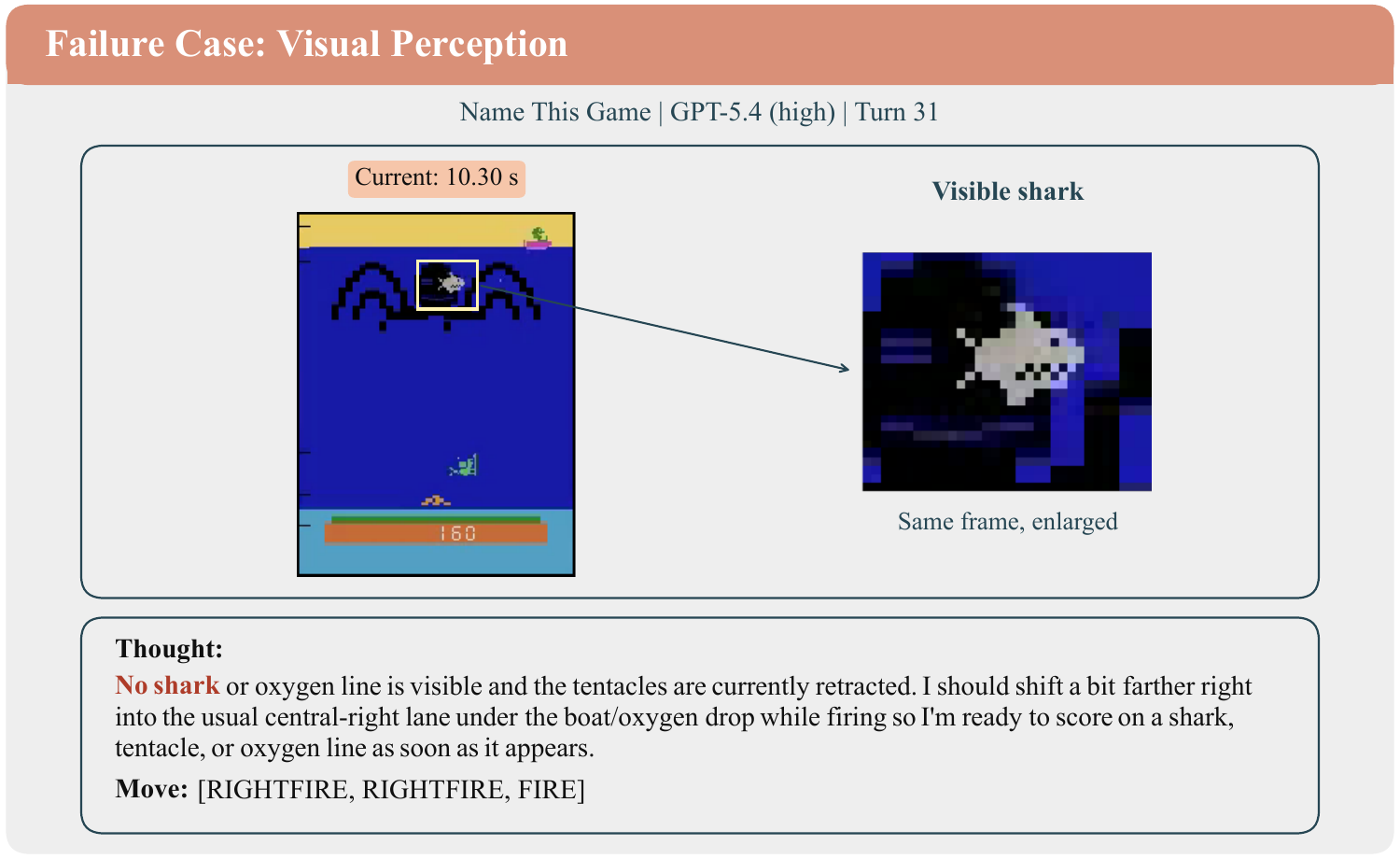}
    \caption{In Name This Game, GPT-5.4 (\texttt{high}) states that no shark is visible despite the white shark in the current frame (10.30\,s; enlarged at right).}
    \label{fig:object_omission_perception}
\end{figure}

\begin{figure}[tbp]
    \centering
    \includegraphics[width=0.9\linewidth]{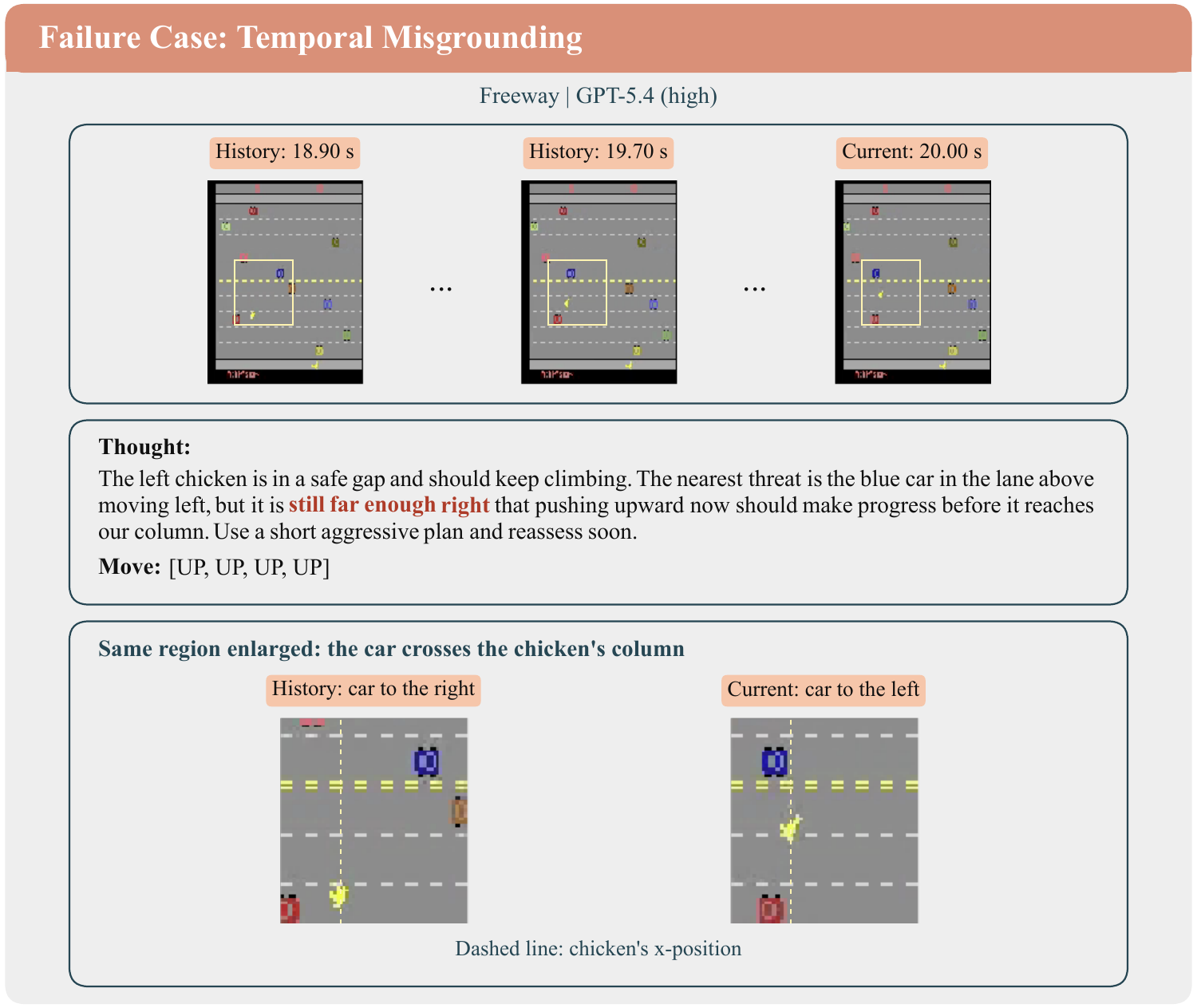}
    \caption{In this Freeway case, GPT-5.4 (\texttt{high}) describes a historical car position as current; with history removed and the remaining input and model settings unchanged, the erroneous relation is no longer stated (see Appendix~\ref{appendix:spatial_history_controls}).}
    \label{fig:temporal_misgrounding}
\end{figure}

\begin{figure}[tbp]
    \centering
    \includegraphics[width=0.9\linewidth]{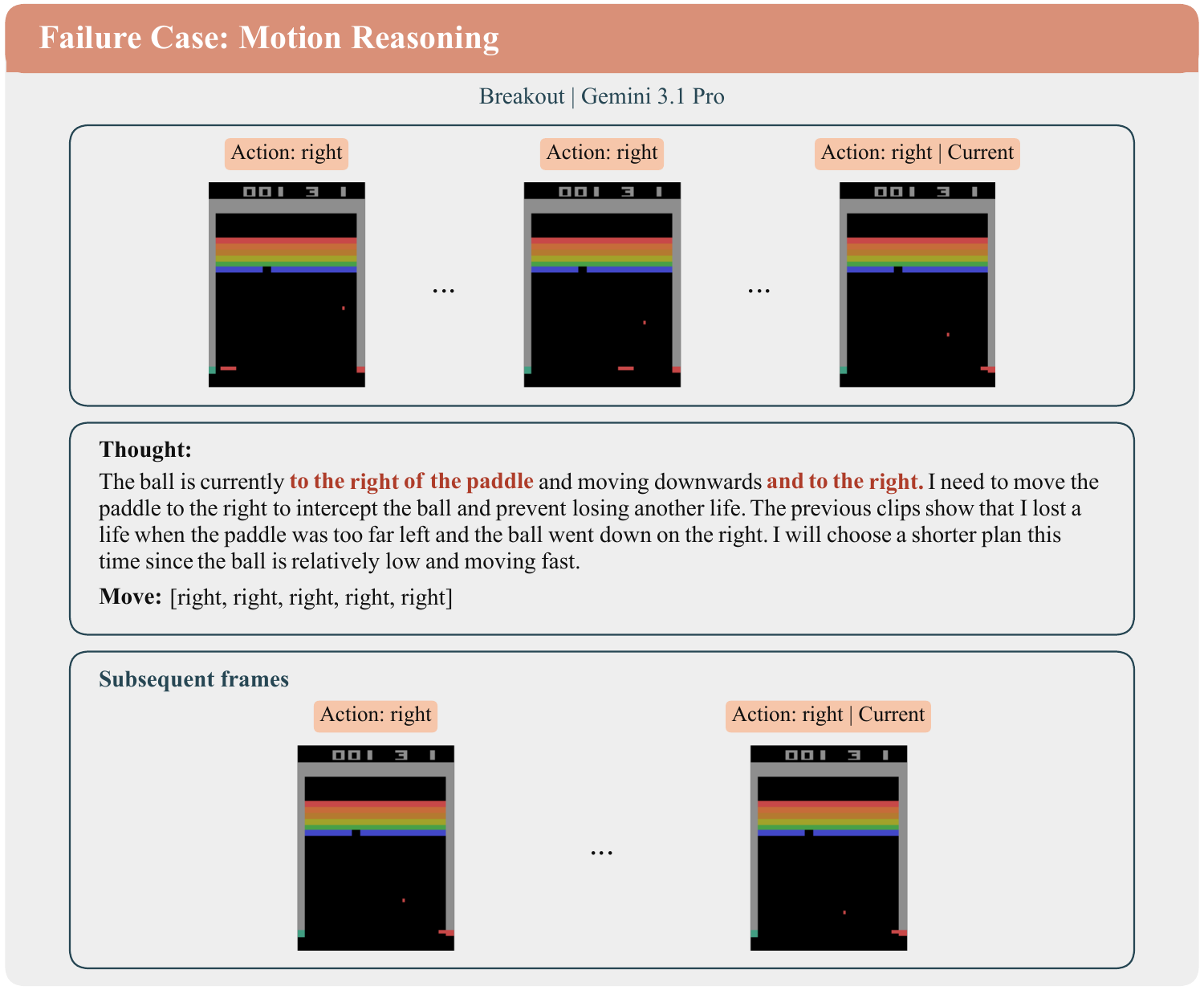}
    \caption{Breakout example of Gemini 3.1 Pro where the model misidentifies the ball’s motion direction.}
    \label{fig:breakout_motion}
\end{figure}

\begin{figure}[tbp]
    \centering
    \includegraphics[width=0.9\linewidth]{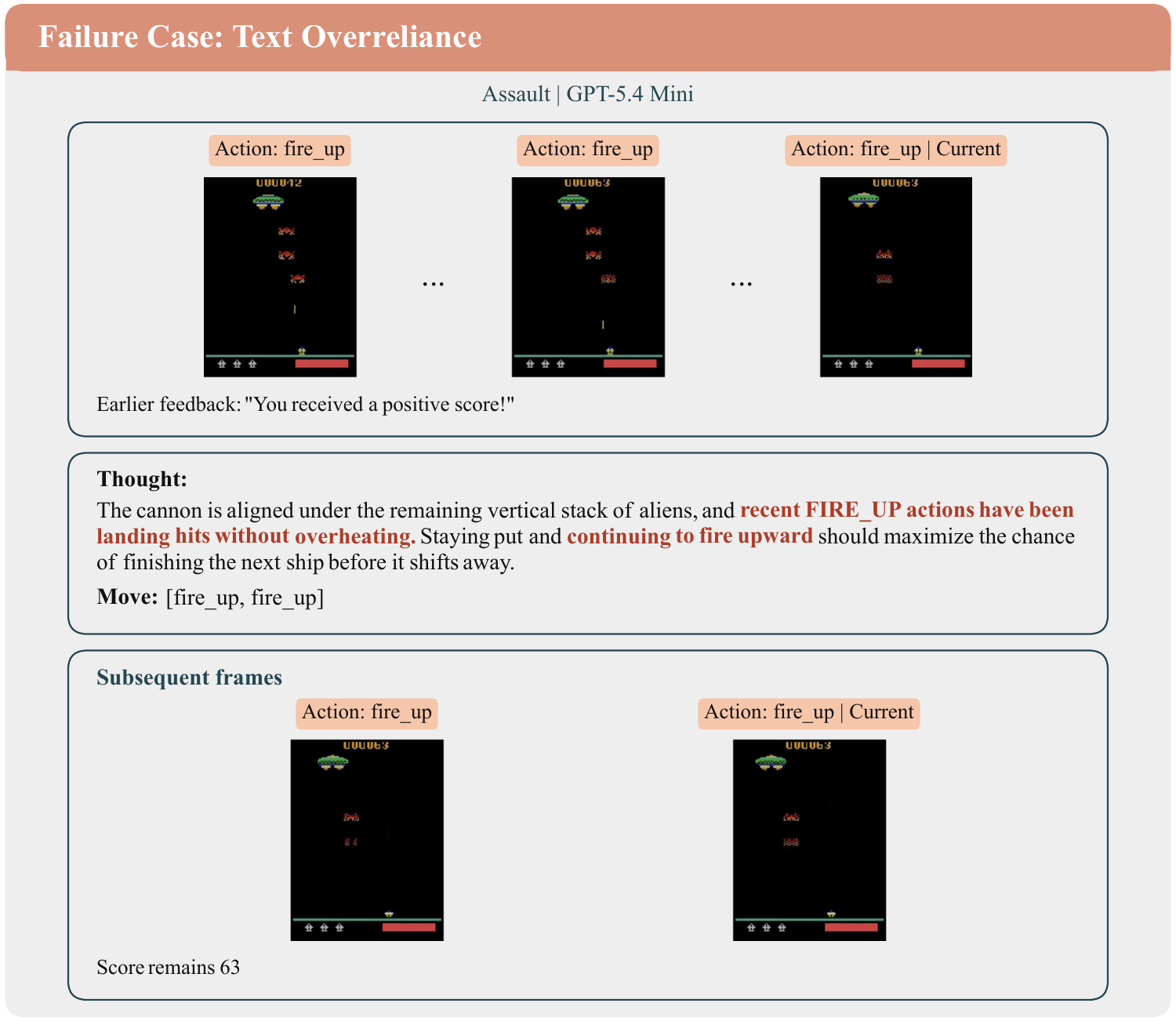}
    \caption{Assault text-history-overreliance case in GPT-5.4 Mini. After an earlier rewarded upward shot, the model says recent shots are hitting despite updated frames showing the remaining alien left of its firing lane. Two more upward shots yield no reward.}
    \label{fig:assault_text_overreliance}
\end{figure}

\begin{figure}[tbp]
    \centering
    \includegraphics[width=0.9\linewidth]{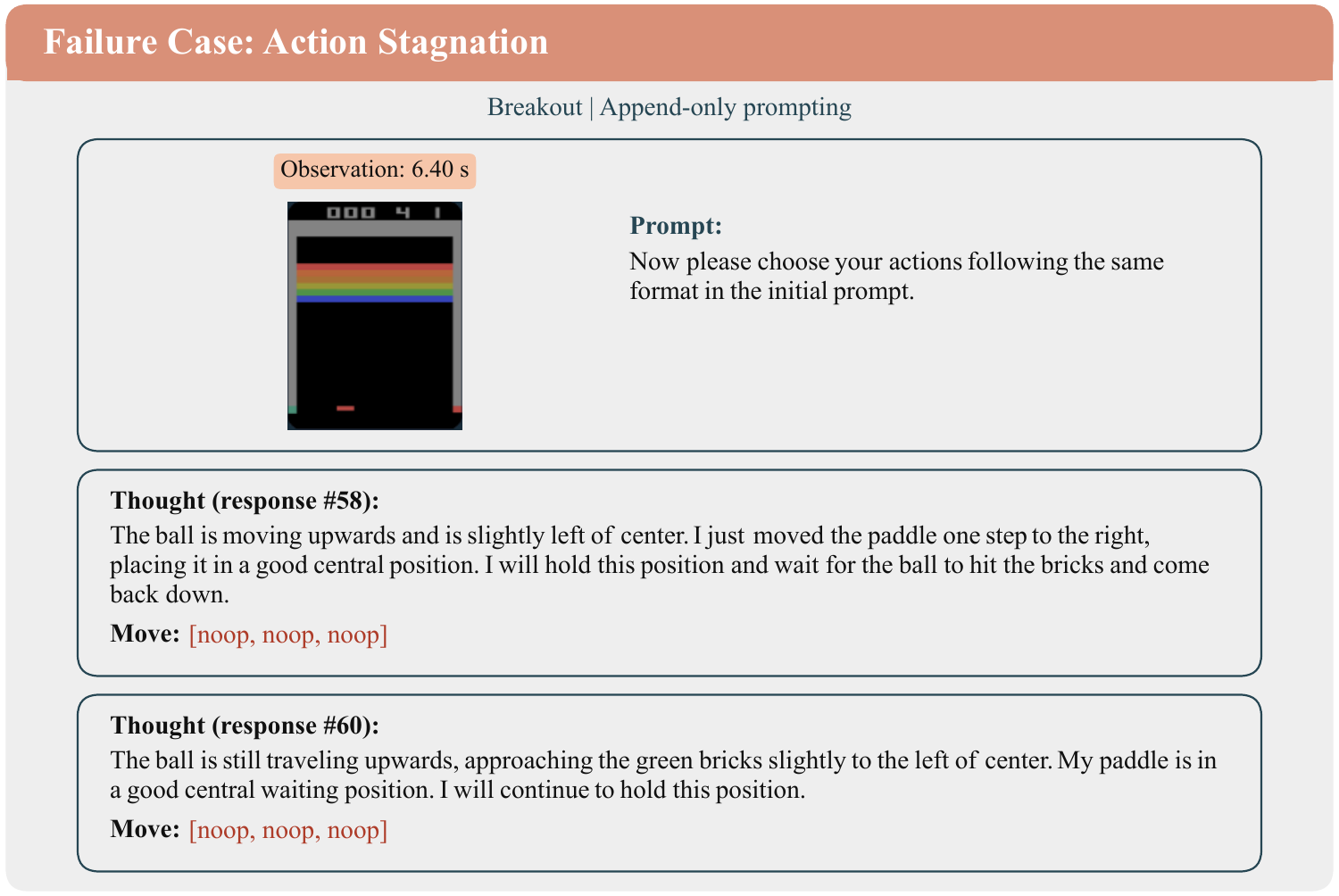}
    \caption{Example of action stagnation under append-only prompting in Breakout. Despite receiving updated frames and feedback, the model repeatedly outputs only \texttt{noop} actions; we observe this behavior in Gemini 2.5 Flash, Gemini 2.5 Pro, and Gemini 3 Flash.}
    \label{fig:append_only_failure_example}
\end{figure}

\begin{itemize}
    \item Fig.~\ref{fig:object_omission_perception} illustrates object omission in Name This Game: GPT-5.4 (\texttt{high}) denies that a shark is visible despite its presence in the current frame. The selected movement and firing actions remain reasonable; the supported error is the stated absence of the shark.
    \item Fig.~\ref{fig:temporal_misgrounding} illustrates a historical-state/current-state mismatch in Freeway: the described car position matches earlier input frames but contradicts the current frame.
    \item Fig.~\ref{fig:breakout_motion} illustrates a motion-reasoning error in Breakout: the relevant objects are visible, but the inferred direction contradicts the ground-truth trajectory.
    \item Fig.~\ref{fig:assault_text_overreliance} illustrates a text-history conflict in Assault: the model repeats a previously rewarded firing action although the target moves out of alignment.
    \item Fig.~\ref{fig:append_only_failure_example} illustrates the long-context failures. Under append-only prompting, the model repeatedly outputs \texttt{noop} actions despite receiving updated frames and feedback. This behavior illustrates how models can fail to update their actions as the game state changes under long-context prompting.
\end{itemize}

\subsection{VLM-Judge Analysis on Gameplay Failures}
\label{appendix:vlm_judge}

The controlled QA diagnostics described in Sec.~\ref{sec:multi_choice_eval} provide a 
breakdown of failures, but they are manually checked and limited in scale. 
Here we provide a VLM-judge analysis of gameplay failures, complementing the controlled QA diagnostics. 

\paragraph{Sampling and adjudication.}
We study Claude Opus 4.6 (\texttt{low}), GPT-5.4 (\texttt{high}), and
Gemini 3.1 Pro (\texttt{low}). For each model--game pair, we consider the
benchmark runs with complete logged trajectories, select the run whose score is
closest to the median, and randomly sample five decision turns at least five
turns apart, yielding 100 turns per model. Turn selection does not condition on
reward or life loss. GPT-5.6-Sol with \texttt{high} reasoning receives the
model's input frames and text, stated rationale, selected actions, and
subsequent observations and feedback. Model identity and reasoning-setting
labels are hidden. The judge first determines whether the evidence supports
an observable gameplay failure, then assigns one primary category:
spatiotemporal reasoning, decision and control, visual perception, exploration,
or task and rule understanding. A penalty alone does not establish a model
error. The judge returns a structured judgment containing failure presence,
one primary category, a subtype, and supporting evidence. A no-observable-failure
label means that the available evidence does not support an error; it does
not establish optimal play. As a sanity check, the judge achieves 93.0\%
accuracy (279/300 responses) on the controlled QA diagnostic set in
Sec.~\ref{sec:multi_choice_eval}.

\begin{samepage}
For example, in Fig.~\ref{fig:breakout_motion}, Gemini 3.1 Pro describes the ball
as moving down-right when the input frames show it moving down-left, then
selects five rightward actions.
Such cases connect an incorrect estimate of the evolving visual state to
the selected control action. Decision and control errors include cases
where a plausible stated plan is not reflected in the executed actions;
visual-perception errors include mistaking the controlled aircraft for an
enemy in Time Pilot.
\par\end{samepage}

\begin{figure}[!htbp]
\centering
\begin{minipage}[c]{0.34\linewidth}
\centering
\includegraphics[width=\linewidth]{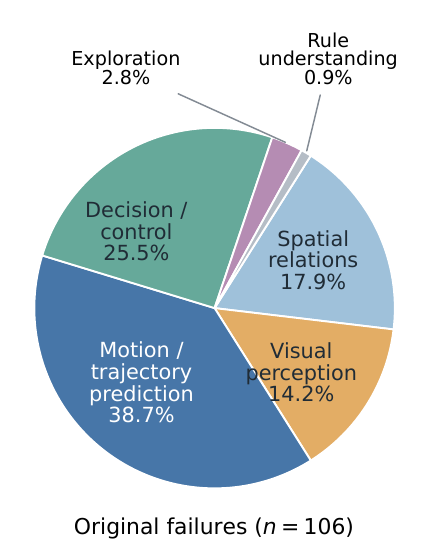}
\end{minipage}\hfill
\begin{minipage}[c]{0.64\linewidth}
\centering
\footnotesize
\textbf{Spatial-relation re-queries ($n=28$)}\\[4pt]
\setlength{\tabcolsep}{3pt}
\renewcommand{\arraystretch}{1.18}
\begin{tabular*}{\linewidth}{@{\extracolsep{\fill}}lrr@{}}
\toprule
Outcome & Complete-history & Current-frame-only \\
\midrule
Still has spatial-relation errors & 11 (39.3\%) & 3 (10.7\%) \\
Correct spatial relations & 17 (60.7\%) & 25 (89.3\%) \\
\bottomrule
\end{tabular*}
\vspace{8pt}

\begin{tabular*}{\linewidth}{@{\extracolsep{\fill}}lr@{}}
\toprule
Paired comparison & Estimate \\
\midrule
Error-rate difference & $-28.6$ percentage points \\
Matched error OR [95\% CI] & 0.20 [0.02, 0.94] \\
Exact McNemar $p$ (two-sided) & 0.0386 \\
\bottomrule
\end{tabular*}
\end{minipage}
\caption{Original VLM-judge failures (left, $n=106$) and paired spatial-relation re-queries (right, $n=28$). Current-frame-only inputs remove history images and associated text while retaining the current frame, game instructions, and model settings. Responses without a spatial-relation statement are re-queried with identical input; the first response stating a relation is assessed, whether correct or incorrect. Correct relations need not explicitly repair the original claim. The matched error odds ratio (OR; current only versus complete input) has an exact 95\% confidence interval.}
\label{fig:vlm_judge_failures}
\end{figure}

\paragraph{Failure categories.}
The judge identifies 106 failures among 300 sampled turns (35.3\%).
Motion/state dynamics and trajectory prediction account for 41 failures
(38.7\%: 32 motion and nine trajectory-prediction errors), decision and control
for 27 (25.5\%), spatial relations for 19 (17.9\%), visual perception for 15
(14.2\%), exploration for three (2.8\%), and task/rule understanding for one
(0.9\%). Fig.~\ref{fig:vlm_judge_failures} retains these original labels and
separately examines 28 spatial-relation cases from the same source runs.

\paragraph{Interpretation and limitations.}
For reporting, we separate spatial relations from motion/state dynamics and
trajectory prediction within the original judge's spatiotemporal category.
Motion/trajectory prediction is the largest reported group, alongside substantial
decision/control, spatial-relation, and perception errors. These results do not establish gameplay
score as a pure measure of spatiotemporal reasoning or measure the causal
effect of each failure on total reward. Attribution remains model-assisted:
the judge can misread visual evidence, and an agent's stated rationale need
not fully explain its action. With only five turns from one selected run per
model--game pair, the results are descriptive and should not be used to rank
models or estimate failure rates across all benchmark trajectories.

\paragraph{Temporal Misgrounding.}
\label{appendix:spatial_history_controls}
To examine whether temporal misgrounding contributes to spatial-relation errors,
we compare complete-history-input and current-frame-only re-queries for spatial-relation failure cases.
Complete input preserves the original prompt and
ordered images; current-frame-only input removes history images, associated
action/reward text, and the history-format explanation, while retaining
the current frame and timestamp, game instructions, action task, and model
settings. We then check whether a spatial relationship is mentioned in the response.
The cases include the 19 judged spatial-relation failures above and 11 additional
failures sampled from the same 60 source runs; excluding two cases that lack a
spatial-relation statement after re-querying leaves 28 cases for analysis.

We assess the current static spatial relations expressed in each
response, including relative position, alignment, distance, and containment.
\textit{Still has a spatial-relation error} means at least one stated relation
is contradicted by the current frame. \textit{Correct spatial relations}
means the stated relations are supported at their expressed level of detail;
it does not require explicitly repairing the originally disputed claim.
Pure object detection, motion direction, and future action predictions are
outside this spatial-only classification. 

As shown in Fig.~\ref{fig:vlm_judge_failures},
complete-history-input re-queries contain spatial errors in 11/28 cases (39.3\%),
compared with 3/28 (10.7\%) for current-frame-only inputs
(two-sided exact McNemar $p=0.0386$). This difference is consistent with
temporal misgrounding as one possible source of spatial-relation errors,
related to reliance on outdated frames~\citep{luo2025thinkingdrifts} and
failures of present-time grounding~\citep{liang2026oasis}.

\subsection{Diagnostic Questions and VLM-Judge Calibration}
\label{appendix:judge_calibration}

\begin{table}[!t]
\centering
\caption{Diagnostic question prompts for each capability group, using Fishing Derby as an example. The long decision prompt additionally prepends the game-specific guidelines.}
\small
\includegraphics[width=0.3\linewidth]{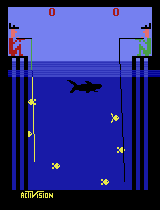}
\vspace{0.5em}

\begin{tabular}{p{0.20\linewidth}p{0.72\linewidth}}
\toprule
\textbf{Capability} & \textbf{Prompt text before interleaved frames} \\
\midrule
Perception: detection &
This is the Atari game Fishing Derby. How many yellow fish are in the game?
Choose from 4, 5, 6, or 7. \\
\addlinespace
Perception: spatial reasoning &
This is the Atari game Fishing Derby. In the last frame, focus on the
deepest fish in the left half of the sea and the left player's fishing line. Is
the left player's fishing line currently to the left or to the right of that
fish? Choose from left or right. \\
\addlinespace
Motion: target object &
This is the Atari game Fishing Derby. Focus on the deepest fish in the
left half of the sea. Which direction is that fish moving? Choose from left or
right. \\
\addlinespace
Motion: controlled object &
This is the Atari game Fishing Derby. Focus on the left player's fishing
line. Across the timestamped sequence, what is the fishing line's overall
movement direction? Choose only one from upright, upleft, downleft, downright,
up, down, left, right, or still. \\
\addlinespace
Decision: short &
This is the Atari game Fishing Derby. Focus on the deepest fish in the
left half of the sea and the left player's fishing line. To catch that deepest
fish, which direction should the left player's fishing line move next? Choose
from left or right. \\
\addlinespace
Decision: long &
This is the Atari game Fishing Derby. \emph{Action Planning Guidelines:
[complete Fishing Derby guide included in the model prompt; omitted here,
approximately 500 tokens of instruction].} Using the frame sequence, focus on
the deepest fish in the left half of the sea and the left player's fishing line.
First identify that target fish and the current position of the left player's
fishing line, then infer their motion. To catch that fish, which direction
should the left player's fishing line move next? Choose from left or right. \\
\bottomrule
\end{tabular}
\label{tab:diagnostic_prompt_templates}
\end{table}

The controlled QA diagnostics in Sec.~\ref{sec:multi_choice_eval} use 60 manually curated question variants across ten games, with five queries per question. Each game contributes two perception questions, two motion questions, and one decision question with short and long prompts. The games are Assault, Boxing, Breakout, Fishing Derby, Freeway, Gopher, Name This Game, Pac-Man, Q*bert, and River Raid. Table~\ref{tab:mini_eval} in the main text reports the diagnostic results.

Table~\ref{tab:diagnostic_prompt_templates} gives the prompt text for the multiple-choice diagnostic protocol using Fishing Derby as an example. These closed-choice questions are distinct from the trajectory-level judge protocol in Appendix~\ref{appendix:vlm_judge}. The remaining samples use the same structure with game-specific object names and choices. Each query begins with the game-identification sentence; frames are provided as an interleaved timestamped sequence; perception questions use their designated single frame; motion questions use a timestamped sequence, while decision questions use the task-specific frame or sequence.

\paragraph{VLM-Judge Calibration.}
On the controlled diagnostic QA set, GPT-5.6-Sol (\texttt{high}), our
trajectory-level judge, achieves 88.0\% perception accuracy, 93.0\% motion
accuracy, and 98.0\% decision accuracy (100 responses per category), with
an overall accuracy of 93.0\% (279/300), clearly higher than the best evaluated
model in Table~\ref{tab:mini_eval} (GPT-5.4 (low), 86.3\%).
For trajectory-level failure
attribution, the judge additionally receives post-action frames and
environment feedback, providing evidence of the selected actions' actual
outcomes.

\subsection{Reasoning Effort and Error Propagation}
\label{appendix:reasoning_propagation}

Fig.~\ref{fig:main} shows that higher reasoning does not consistently improve performance. We investigate whether failures arise from state estimation, planning, or both. For state-estimation failures, we further define a failure as \emph{propagated} when an incorrect state estimate leads to an incorrect or clearly suboptimal action.

We use the VLM judge to identify failures and randomly select 50 failed turns
each for GPT-5.4 (high) and GPT-5.4 (low) for further analysis. We ask the judge to classify each failure as a spatiotemporal state-estimation error, a planning error despite a substantially correct state estimate, or another type of failure. For state-estimation errors, we assess whether the incorrect estimate leads to an erroneous action. An incorrect description alone does not count as propagation if the selected action remains reasonable or the link to an action error is unclear.

\begin{table}[!htbp]
\centering
\small
\caption{Failure types among 50 failed turns per setting. State estimation, planning with a substantially correct state estimate, and other failures are mutually exclusive. Propagation is measured only among state-estimation failures.}
\label{tab:reasoning_propagation}
\begin{tabular}{@{}lrrrr@{}}
\toprule
Setting & \shortstack{State-estimation\\failures} & \shortstack{Planning failures\\with correct state} & Other & Propagation \\
\midrule
GPT-5.4 (high) & 37/50 (74\%) & 3/50 (6\%) & 10/50 (20\%) & 26/37 (70.3\%) \\
GPT-5.4 (low) & 29/50 (58\%) & 12/50 (24\%) & 9/50 (18\%) & 20/29 (69.0\%) \\
\bottomrule
\end{tabular}
\end{table}

Spatiotemporal state estimation accounts for 66/100 failures, compared with
15/100 attributed to planning despite a substantially correct state estimate.
The remaining 19 cases involve static visual errors or other bottlenecks.
Among state-estimation failures, 26/37 (70.3\%) in the high-effort setting
and 20/29 (69.0\%) in the low-effort setting propagate into erroneous actions,
for a combined 46/66 (69.7\%). One case in each setting has insufficient
evidence to establish propagation and is excluded from the numerator.

Under high reasoning effort, the failure composition shifts toward state estimation (74\% versus 58\% of sampled failures), while planning accounts for a smaller share (6\% versus 24\%). Because only failed turns are sampled, these shares describe the composition of failures rather than absolute failure rates. In both settings, state-estimation errors propagate into incorrect actions at a similar rate (approximately 70\%), suggesting that higher reasoning effort does not reduce the downstream impact of spatiotemporal misestimation.

This is consistent with prior findings: VideoAuto-R1~\cite{videoauto-r1} finds that direct answering often matches or outperforms chain-of-thought on perception-oriented video tasks, while More Thinking, Less Seeing?~\cite{more-thinking-less-seeing} shows that longer reasoning chains can reduce attention to visual inputs and amplify visual hallucinations.

\subsection{Text-History Overreliance}
\label{appendix:text_history_audit}

We use the VLM judge to identify failures involving reliance on
stale textual history in 240 randomly sampled turns across all 20 games and
four model settings: GPT-5.4 (high), Gemini 3.1 Pro (low),
GPT-5.4 Mini (low), and Gemini 2.5 Flash (off). We treat GPT-5.4 (high) and Gemini 3.1 Pro (low) as the stronger settings and GPT-5.4 Mini (low) and Gemini 2.5 Flash (off) as the weaker settings. Each setting
contributes three turns per game, for 60 turns in total, sampled independently
of reward or life loss. We label a failure as
text-history overreliance when the model follows a past textual cue that
conflicts with current visual evidence instead of updating its decision
from the frames. Repeated actions or an incorrect motion estimate alone
do not satisfy this definition.

\begin{table}[!htbp]
\centering
\small
\caption{Text-history overreliance in sampled gameplay turns.
Failures counts the observable failures identified by the judge, and
Overreliance Rate is the percentage of these failures attributed to
text-history overreliance.}
\label{tab:text_history_audit}
\begin{tabular}{@{}lrrrr@{}}
\toprule
Model setting & Turns & Failures & \shortstack{Text Overreliance\\Failures} & \shortstack{Overreliance\\Rate} \\
\midrule
GPT-5.4 (high) & 60 & 21 & 2 & 9.5\% \\
Gemini 3.1 Pro (low) & 60 & 20 & 0 & 0.0\% \\
GPT-5.4 Mini (low) & 60 & 24 & 5 & 20.8\% \\
Gemini 2.5 Flash (off) & 60 & 43 & 10 & 23.3\% \\
\midrule
Two stronger settings & 120 & 41 & 2 & 4.9\% \\
Two weaker settings & 120 & 67 & 15 & 22.4\% \\
\bottomrule
\end{tabular}
\end{table}

Conditional on observable failure, text-history overreliance occurs in
4.9\% of cases for the two stronger settings and 22.4\% for the two weaker
settings (two-sided Fisher exact test, $p=0.0154$). Measured per sampled
turn, the rates are 1.7\% (2/120) and 12.5\% (15/120), respectively
($p=0.0016$).
These results suggest that weaker models are more prone to relying on stale
textual history, while stronger models exhibit this problem less frequently.

\section{Cost Analysis}
\label{appendix:cost}

AtariBench does not require model training or local GPU inference in the released harness. The local worker runs ALE simulation, frame capture, prompt assembly, parsing, and artifact writing on CPU; the dominant reproducibility cost is paid VLM API usage. A rollout issues one model query per closed-loop decision turn, so the number of calls depends on the model's chosen action sequence length rather than on the raw Atari frame count.

We calculate Gemini 3.1 Pro evaluation cost from a measured AtariBench token footprint, covering 20 games with 10 runs per game. The token accounting includes input, output, and thinking-token components, and we price this 200-run closed-loop visual-control footprint with the Gemini 3.1 Pro token prices used for the paper.

\begin{table}[!htbp]
\centering
\caption{Estimated Gemini 3.1 Pro API cost for 200 AtariBench runs (20 games $\times$ 10 runs).}
\label{tab:gemini_cost}
\small
\begin{tabular}{@{}lrrrrrr@{}}
\toprule
\textbf{Setting} & \textbf{Runs} & \textbf{Calls} & \textbf{Input tok.} & \textbf{Output tok.} & \textbf{Thinking tok.} & \textbf{Est. cost} \\
\midrule
Ours & 200 & 27{,}938 & 140.81M & 2.85M & 17.18M & \$522 \\
\bottomrule
\end{tabular}
\end{table}

The estimate uses the Gemini 3.1 Pro prices for prompts under 200k tokens: \$2 per million input tokens and \$12 per million output tokens, with thinking tokens counted at the output rate. The resulting formula is
\[
\mathrm{cost} = \frac{2N_{\mathrm{input}} + 12(N_{\mathrm{output}} + N_{\mathrm{think}})}{10^6}.
\]
The matched setting has an average of 5.0k input tokens per model call under the structured-history memory mode, so the under-200k prompt tier is the appropriate estimate. Context caching is not included in these totals; when cache accounting is used, the price would be lower.

\section{Ethics and Release}
\label{appendix:ethics_release}

AtariBench uses the Arcade Learning Environment (ALE) as an external
GPL-2.0-licensed dependency. We use ALE version 0.8.1 (\texttt{ale-py})
for evaluation and preserve the applicable copyright and license notices.
The required Atari ROMs are obtained through \texttt{ale-py}
following ALE's installation instructions.\footnote{\url{https://github.com/Farama-Foundation/Arcade-Learning-Environment}}
Users should follow the applicable dependency and game-asset terms.

The AtariBench code is released as a self-contained Colab notebook, linked
from the project website (\url{https://mignonjia.github.io/AtariBench/}),
under the Apache License 2.0. The notebook contains the full evaluation
harness (game prompts, prompt builder, response parser, and gameplay loop),
a Gemini client, and the settings needed to reproduce the paper's Gemini
experiments. Other providers can be evaluated by supplying a different
model client to the same harness. For illustration,
the project website shows a few short gameplay videos from evaluated
runs. We will not bundle Atari ROM binaries,
rendered-frame datasets, or bulk visual trajectories with the benchmark
release. Users can generate gameplay trajectories locally through the ALE
dependency.

\end{document}